\documentclass[journal]{IEEEtran}
\ifCLASSINFOpdf
\else
\fi

\usepackage{hyperref}
\usepackage{float}
\usepackage{ulem}
\usepackage{booktabs}
\usepackage{amssymb}
\usepackage{graphicx}
\usepackage{indentfirst}
\usepackage{subfigure}
\usepackage{algorithm}
\usepackage{algorithmic}
\usepackage{amsmath}
\usepackage{multirow}
\usepackage{threeparttable}  
\usepackage{url}
\usepackage{bibentry}
\usepackage{epsfig}
\usepackage{color}
\usepackage{epstopdf}
\usepackage{rotating}
\usepackage{caption}
\usepackage{multirow}
\usepackage{stfloats} 
\usepackage{color}% This package is for using color package.
\usepackage{bm}
\usepackage{graphicx}
\usepackage{ntheorem}
\usepackage{fancybox}
\usepackage{tikz}
\usetikzlibrary{calc,3d}
\usepackage{colortbl}  %彩色表格需要加载的宏包
\usepackage{xcolor}
\newtheorem{definition}{Definition}

\begin{document}
    %
    % paper title
    % Titles are generally capitalized except for words such as a, an, and, as,
    % at, but, by, for, in, nor, of, on, or, the, to and up, which are usually
    % not capitalized unless they are the first or last word of the title.
    % Linebreaks \\ can be used within to get better formatting as desired.
    % Do not put math or special symbols in the title.
    %\title{Robust Vertical Federated Learning with Tensor Decomposition for Missing Features }
    
    %\title{Privacy-Preserving Vertical Federated Learning with Tensor Decomposition for Missing Features}
    %\title{Robust Vertical Federated Learning with Tensor Decomposition for Data Missing Features}
    % \title{Advances in Robust Federated Learning: Heterogeneity Considerations}
    \title{Advances in Robust Federated Learning: A Survey with Heterogeneity Considerations}

    \author{
        Chuan Chen \textit{Member, IEEE}, Tianchi Liao, Xiaojun Deng, Zihou Wu, Sheng Huang, Zibin Zheng* \textit{Fellow, IEEE}
        % <-this % stops a space
        \thanks{
            Chuan Chen, Xiaojun Deng, and Zihou Wu were with the School of Computer Science and Engineering, Sun Yat-sen University, Guangzhou 510006, China. 
            Chuan Chen was also with the Guangdong Key Laboratory of Big Data Analysis and Processing
            (e-mail: 
            chenchuan@mail.sysu.edu.cn, dengxj26@mail2.sysu.edu.cn, wuzh78@mail2.sysu.edu.cn)
            
            Tianchi Liao, and Zibin Zheng were with the School of Software Engineering, Sun Yat-sen University, Zhuhai 519000, China. (e-mail: liaotch@mail2.sysu.edu.cn, zhzibin@mail.sysu.edu.cn).
            Zibin Zheng was also with the Zhuhai Key Laboratory of Trusted Large Language Models.
            
            Sheng Huang was with the School of Systems Science and Engineering, Sun Yat-sen University, Guangzhou 510006, China. (e-mail: huangsh253@mail2.sysu.edu.cn).
            % Chuan Chen is the corresponding author.
            
            The research is supported by the National Key Research and Development Program of China (2023YFB2703700), the National Natural Science Foundation of China (62176269), the Guangzhou Science and Technology Program (2023A04J0314).
        }
        
        %	In the meanwhile, they study in National Engineering Research Center of Digital Life, Sun Yat-sen University, Guangzhou, China (Corresponding author: Chuan Chen. E-mail:liaotch@mail2.sysu.edu.cn, yangjinghua110@126.com, wuzhb6@mail2.sysu.edu.cn,  chenchuan@mail.sysu.edu.cn, zhzibin@mail.sysu.edu.cn).}% <-this % stops a space
    % <-this % stops a space
    %\thanks{J. Doe and J. Doe are with Anonymous University.}% <-this % stops a space
    %\thanks{Manuscript received April 19, 2005; revised August 26, 2015.}
    
        \thanks{Zibin Zheng is the corresponding author.}
}

% make the title area
\maketitle

% As a general rule, do not put math, special symbols or citations
% in the abstract
\begin{abstract}
    In the field of heterogeneous federated learning (FL), the key challenge is to efficiently and collaboratively train models across multiple clients with different data distributions, model structures, task objectives, computational capabilities, and communication resources. This diversity leads to significant heterogeneity, which increases the complexity of model training. In this paper, we first outline the basic concepts of heterogeneous FL and summarize the research challenges in FL in terms of five aspects: data, model, task, device and communication. In addition, we explore how existing state-of-the-art approaches cope with the heterogeneity of FL, and categorize and review these approaches at three different levels: data-level, model-level, and architecture-level. Subsequently, the paper extensively discusses privacy-preserving strategies in heterogeneous FL environments. Finally, the paper discusses current open issues and directions for future research, aiming to promote the further development of heterogeneous FL.
    
\end{abstract}


% For peer review papers, you can put extra information on the cover
% page as needed:
% \ifCLASSOPTIONpeerreview
% \begin{center} \bfseries EDICS Category: 3-BBND \end{center}
% \fi
%
% For peerreview papers, this IEEEtran command inserts a page break and
% creates the second title. It will be ignored for other modes.
\IEEEpeerreviewmaketitle

\section{Introduction}
With the popularization of mobile devices and 5G networks, a large amount of data is generated on edge devices every day\cite{li2021survey}. Traditional machine learning (ML) methods require centralized data storage and processing for model training, which poses significant challenge to data transmission and storage. Moreover, with the frequent occurrence of data leakage, the issue of privacy protection attracts more and more attention. 
Thus, institutions and enterprises are looking for a new ML paradigm, which can utilize distributed data resources for training while protecting data privacy and security.

In 2016, Google researchers proposed the concept of federated learning (FL) \cite{mcmahan2017communication}. The core idea about FL is to train and update models locally on data owners (i.e. clients) and aggregates model updates on a third party (i.e. server), without the need to transfer the raw data to a central server. By this way, FL can significantly reduce data transmission in the network, thus reducing the possibility of privacy leakage and the requirement for centralized computing and storage resources. FL not only presents the respect and protection for privacy, but also demonstrates how to maximize the utility of data while ensuring privacy in the era of big data. At present, FL is becoming an effective way for balancing the utility and privacy of data, bringing new development directions and opportunities to ML. FL has demonstrated its broad application potential in fields such as intelligent healthcare \cite{antunes2022federated}, finance \cite{nevrataki2023survey}, the Internet of Things \cite{nguyen2021federated}, and smart cities \cite{zheng2022applications}.

However, in the application scenarios of typical FL, there are multiple heterogeneity issues among clients, which brings challenges to collaboratively model training. The presence of heterogeneity not only significantly damages the generalization ability and prediction accuracy of trained model, but also to some extent restrict the application of FL. Specifically, FL often counters the following heterogeneity:
% \textbf{Data heterogeneity}: The clients often far away from each other geographically and lack communications, and each client may collect data from different data source, causing skews in the distributions of data collected by clients \cite{wang2020optimizing}. Moreover, the method of sampling and processing data and the amount of collected data may also vary across different clients, resulting in quality skew and quantity skew.
    \textbf{Data heterogeneity} includes several types of skews in the data collected by clients. Before FL, each client may collect data from different data source, which may result in distribution skew, label skew and feature skew \cite{wang2020optimizing,mai2024fgtl}. Moreover, the method of sampling and processing data and the amount of collected data may also vary across different clients, resulting in quality skew and quantity skew.
%\textbf{Model heterogeneity}: Due to device limitations, task types and other reasons, the clients have to customize their own model structures and jointly train models in a federated manner, which is model structure heterogeneity \cite{li2019fedmd}. Moreover, the adopted optimizer of different clients may also be different.
    \textbf{Model heterogeneity} includes model structure heterogeneity and model optimizer heterogeneity. Model structure heterogeneity means that the clients have to customize their own model structures and jointly train models in a federated manner, due to device limitations, task types and other reasons \cite{li2019fedmd}. Model optimizer heterogeneity means that the adopted optimizer of different clients may also be different.
% \textbf{Task heterogeneity}: Sometimes different clients need to train models for different tasks. 
    \textbf{Task heterogeneity} means that different clients need to train models for different tasks. For example, some clients may need to train models for image classification, while some other clients may need to train models for semantic segmentation \cite{smith2017federated}. 
%\textbf{Communication heterogeneity}: In typical FL, each round of FL requires transmission of model parameters or gradients between the server and the clients. But in reality, the participants of FL are usually under different network environments, and communication bandwidth and budgets of some participants are limited, which is called communication heterogeneity \cite{abad2020hierarchical}.
    \textbf{Communication heterogeneity} means that the participants of FL are usually under different network environments, and communication bandwidth and budgets of some participants are limited \cite{abad2020hierarchical}. Therefore, limited bandwidth and unstable connection can significantly reduce the efficiency of FL.
%\textbf{Device heterogeneity}: The typical FL algorithms usually require local training on the clients. But due to limited computing power, storage resources, energy supply, etc., the clients may update local models inconsistently and thus lower the efficiency of the whole system \cite{pfeiffer2023federated}.
    \textbf{Device heterogeneity} means that the clients vary in device performance, including computing power, storage resources, energy supply, etc. Since typical FL algorithms usually require several epochs of local training, the clients may update local models inconsistently and thus lower the efficiency of the whole system \cite{pfeiffer2023federated}.

Although FL is originally proposed for helping protect privacy of the clients, the risk of privacy leakage has not been completely eliminated. Besides heterogeneity, FL is still subject to some privacy attacks. The reason is that attackers can easily access client uploads to analyze privacy information. For example, the malicious attackers may intercept the uploaded parameters or gradients and reconstruct the client’s local data by Gradient Inversion \cite{zhu2019deep} or Model Inversion \cite{geiping2020inverting}. Moreover, attackers may also use Membership Inference Attack  \cite{nasr2019comprehensive} to determine whether certain data is includes in local training, or use Property Inference Attack to infer private information of some property, like the data sampling environment and label distribution \cite{lyu2022privacy}. Thus, it is necessary to employ some privacy-preserving mechanisms in FL to enable clients to communicate with servers and other clients in a more secure way.

% Research addressing heterogeneous FL is currently generating a lot of buzz, and we investigate a number of frequently cited or recent surveys related to heterogeneity and privacy issues.
Research addressing heterogeneous FL is currently generating a lot of buzz, and we investigate some highly cited and recent surveys related to heterogeneity and privacy. {While they all provide detailed descriptions of their respective topics of focus, compared to our work, they overlook some less researched but equally important issues and techniques, and most of them cannot sufficiently cover both heterogeneity and privacy. }
% While they all provide detailed descriptions of their respective topics of focus, they omit some of the lesser mentioned issues and types of technology.
% Although they have all provided detailed descriptions and abundant related researches for their respective focused topics, they still overlook some less mentioned types of issues and techniques. 
{In this paper, we first provide a detailed categorization of several types of heterogeneity and discuss the heterogeneity challenges that may be encountered when applying FL in real-world environments as well as state-of-the-art solutions. Considering that privacy preservation is one of the core requirements of FL, we discuss the application of existing privacy-preserving techniques to FL heterogeneous scenarios, exploring the trade-off relationship between privacy and utility, and presenting the latest research progress in related areas. }
% In this paper, we first categorize several types of heterogeneity in detail, and explore heterogeneity challenges may encountered while applying FL for practical environments, as well as solutions. For each type of heterogeneity, we not only present detailed and formal description, but also conduct extensive survey for comprehensively summarizing the state-of-the-art solutions. Moreover, given that privacy protection is one of the core demands of FL, we also focus on privacy protection problems and solutions while applying FL in scenarios with heterogeneity. We discuss the application of existing privacy protection techniques in FL, delving into the trade-off between privacy and utility, and introduce the advanced researches in relevant fields. 
We list the differences between our survey and the above other surveys in \autoref{tab:comparison}.
Thus, the main contributions of this work are threefold: 

\begin{table*}[t] %voc table result
    \centering
    \vspace{-0.6cm}
    \caption{Comparison between other surveys and ours}
    \resizebox{\textwidth}{!}{
        \begin{tabular}{c|c|c|c|c|c|c|c|c|c|c|c|c}
            \toprule
            \multicolumn{2}{c|}{\multirow{2}{*}{{\textbf{Types of Heterogeneity}}}}
            & \multicolumn{10}{c|}{{\textbf{Comparison of Surveys}}}
            & \multirow{2}{*}{{\textbf{Representative Methods}}} \\ \cline{3-12}
            \multicolumn{2}{c|}{} & \cite{ye2023heterogeneous}
            & \cite{wen2023survey}
            & \cite{zhang2021survey}
            & \cite{ji2024emerging}
            & \cite{pfeiffer2023federated}
            & \cite{gao2022survey}
            & \cite{mengistu2024survey} 
            & \cite{guendouzi2023systematic} 
            & \cite{lu2024federated} 
            & Ours \\ \hline
            %\multicolumn{2}{c|}{\textbf{Focus}} 
            %& FL in IoT & FL in IoT & FL in IoT & FL in IoT & FL in IoT & FL in IoT & FL in IoT & FL in IoT & FL in IoT \\ \hline
%				\multirow{5}{*}{\textbf{Data Heterogeneity} } 
            \multirow{5}{*}{\begin{tabular}[c]{c}\textbf{Data Heterogeneity}\\ \textbf{Fig. \ref{fig:DH}}\end{tabular}}
            & Distribution Skew & $\times$ & $\checkmark$ & $\checkmark$ & $\checkmark$ & $\checkmark$ & $\checkmark$ & $\checkmark$ & $\checkmark$ & $\checkmark$ & $\checkmark$ 
            & {\cite{li2021model}, \cite{luo2021no}, \cite{xin2020private}, \cite{xu2023personalized}, \cite{zhang2022fine}, \cite{chen2024spectral}, \cite{smith2017federated}, \cite{chen2018federated}, \cite{li2020federated}, \cite{karimireddy2020scaffold}, \cite{zhi2024knowledge}, \cite{lu2023auction}, \cite{sattler2020cfl}, \cite{lu2023auction}}\\ \cline{2-13}
            & Label Skew & $\checkmark$ & $\times$ & $\times$ & $\times$ & $\times$ & $\checkmark$ & $\times$ & $\times$ & $\checkmark$ & $\checkmark$ 
            & {\cite{collins2021exploiting}, \cite{luo2021no}, \cite{kim2022multi}} \\ \cline{2-13}
            & Feature Skew & $\checkmark$ & $\times$ & $\times$ & $\times$ & $\times$ & $\checkmark$ & $\checkmark$ & $\checkmark$ & $\checkmark$ & $\checkmark$ & {\cite{liu2020secure}, \cite{wu2022practical}, \cite{huang2022learn}, \cite{shen2022cd2}, \cite{huang2022learn}, \cite{jeong2022factorized}} \\ \cline{2-13}
            & Quality Skew & $\checkmark$ & $\times$ & $\times$ & $\times$ & $\times$ & $\times$ & $\checkmark$ & $\times$ & $\times$ & $\checkmark$ & {\cite{li2024feddiv}, \cite{fang2022robust}, \cite{lu2024federated}} \\ \cline{2-13}
            & Quantity Skew & $\checkmark$ & $\times$ & $\times$ & $\times$ & $\checkmark$ & $\times$ & $\checkmark$ & $\times$ & $\times$ & $\checkmark$ 
            & {\cite{ribero2022federated}} \\ \hline
%				\multirow{2}{*}{\textbf{Model Heterogeneity}}
            \multirow{2}{*}{\begin{tabular}[c]{c}\textbf{Model Heterogeneity}\\ \textbf{Fig. \ref{fig:MH}}\end{tabular}}
            & Structure Heterogeneity & $\checkmark$ & $\checkmark$ & $\checkmark$ & $\checkmark$ & $\checkmark$ & $\checkmark$ & $\checkmark$ & $\checkmark$ & $\times$ & $\checkmark$ & {\cite{li2019fedmd}, \cite{liang2020think}, \cite{wu2022communication}, \cite{tan2022fedproto}, \cite{yi2023fedgh}, \cite{shen2020federated}, \cite{huang2022learn}, \cite{wang2024towards}} \\ \cline{2-13}
            & Optimizer Heterogeneity & $\times$ & $\times$ & $\times$ & $\times$ & $\times$ & $\times$ & $\times$ & $\times$ & $\times$ & $\checkmark$ & {\cite{nguyen2022federated}} \\ \hline
            \multicolumn{2}{c|}{\textbf{Task Heterogeneity Fig. \ref{fig:TH}}} & $\times$ & $\times$ & $\times$ & $\times$ & $\checkmark$ & $\times$ & $\times$ & $\times$ & $\checkmark$ & $\checkmark$ & {\cite{smith2017federated}, \cite{arivazhagan2019federated}, \cite{liang2020think}}\\ \hline
            \multicolumn{2}{c|}{\textbf{Communication Heterogeneity} \textbf{Fig. \ref{fig:CH}}} & $\checkmark$ & $\times$ & $\checkmark$ & $\times$ & $\checkmark$ & $\checkmark$ & $\checkmark$ & $\checkmark$ & $\checkmark$ & $\checkmark$ & {\cite{wu2022practical}, \cite{li2021fedmask}, \cite{tan2022fedproto}, \cite{diao2020heterofl}, \cite{pfeiffer2024aggregating}, \cite{sattler2019robust}, \cite{shah2021model}, \cite{wu2022communication}, \cite{deng2024communication}, \cite{ye2022decentralized}} \\ \hline
            \multicolumn{2}{c|}{\textbf{Device Heterogeneity} \textbf{Fig. \ref{fig:DevH}}}& $\checkmark$ & $\times$ & $\checkmark$ & $\times$ & $\checkmark$ & $\checkmark$ & $\checkmark$ & $\checkmark$ & $\checkmark$ & $\checkmark$ & {\cite{li2020federated}, \cite{diao2020heterofl}, \cite{li2021fedmask}, \cite{yi2024fedp3}, \cite{pfeiffer2024aggregating}, \cite{yao2021fedhm}, \cite{chen2024fed}, \cite{he2020group}, \cite{xu2023asynchronous}} \\ \hline
%				\multirow{4}{*}{\textbf{Privacy Mechanism Fig. \ref{fig:P}}}
            \multirow{4}{*}{\begin{tabular}[c]{c}\textbf{Privacy Mechanism}\\ \textbf{Fig. \ref{fig:P}}\end{tabular}}	
            & Differential Privacy & $\checkmark$ & $\checkmark$ & $\checkmark$ & $\checkmark$ & $\times$ & $\checkmark$ & $\checkmark$ & $\checkmark$ & $\checkmark$ & $\checkmark$ & {\cite{mcmahan2017learning}, \cite{wei2020federated}, \cite{seif2020wireless}, \cite{liu2021projected}, \cite{noble2022differentially}} \\ \cline{2-13}
            & Homomorphic encryption & $\times$ & $\checkmark$ & $\checkmark$ & $\checkmark$ & $\times$ & $\checkmark$ & $\checkmark$ & $\checkmark$ & $\checkmark$ & $\checkmark$ & {\cite{aono2017privacy}, \cite{xu2019hybridalpha}} \\ \cline{2-13}
            & Model Watermark & $\times$ & $\times$ & $\times$ & $\times$ & $\times$ & $\times$ & $\times$ & $\times$ & $\times$ & $\checkmark$ & {\cite{tekgul2021waffle}, \cite{nie2024fedcrmw}, \cite{han2021application}} \\ \cline{2-13}
            & Blockchain & $\times$ & $\times$ & $\times$ & $\times$ & $\times$ & $\times$ & $\times$ & $\checkmark$ & $\checkmark$ & $\checkmark$ & {\cite{cui2020creat}, \cite{li2020blockchain}, \cite{li2023veryfl}} \\ \bottomrule
    \end{tabular}}
    \label{tab:comparison}
\end{table*}

\begin{enumerate}
    
%		\item We are 
%		\item We utilize 
    
    \item We systematically categorize and analyze key components in heterogeneous FL, including data, model, task, communication, and device heterogeneity. For each type, we discuss in detail the algorithmic challenges it raises. {Comparing to previous surveys, we further refine some types of heterogeneity. For example, we further divide skews happened in label space into distribution skew and label skew, which are similar but may have different impact on the effect of FL.}
    
    \item {Considering that some techniques can be applied to different types of heterogeneity, we discuss the FL methods in depth based on their adopted techniques, and categorize existing state-of-the-art methods into data-level, model-level, and architecture-level methods, and emphasize the types of heterogeneity targeted by each method.}
    
    \item We summarize the current privacy-preserving strategies in heterogeneous FL and explore possible future research directions with a view to promoting a safer and more efficient development of the field. {Specifically, we discuss some privacy-preserving techniques that can be easily affected by heterogeneity as well as the corresponding countermeasures, which significantly matters but are not often explored in previous survey works.}
\end{enumerate}

\section{Classification and Challenges in Heterogeneous Federated Learning}
% In this section, we first provide a typical FL formulation to illustrate the typical FL process through an example. 
In this section, we illustrate the typical FL process through the equation.
FL is ML setup in which multiple clients collaboratively train a model while protecting data privacy \cite{mcmahan2017learning}. 
% Each participant trains the model on local data and then sends only updates to the model (e.g., weights or gradients) to a central server. The server aggregates these updates and sends the aggregated results back to the participants to update their local models in the next iteration. In this way, federated training produces a global model that performs well on the entire dataset without sharing any raw data. 
In a classic FL framework \cite{mcmahan2017communication}, assuming that there are $K$ clients participating in the training, for $k$-th client $C_k$ has a private dataset $D_k=\left\lbrace (x_i^k, y_i^k) \right\rbrace_{i=1}^{N_k} $ that satisfies $|x^k|=N_k$ and $N=\sum_{k=1}^{K}N_k$. The local distribution of the data of client $k$ is denoted as $P_k(x,y)$, where $x$ and $y$ denote the features of the data samples and the labels.
Specifically, FL can be formally defined as the following optimization problem:
\begin{equation}
    \label{eqb:obj}
    \min_{\theta} F(\theta) = \sum_{k=1}^{K}p_k F_k(\theta)
\end{equation}
where $\theta$ is the global model parameter. $F_k(\theta)$ is the local loss function on client $k$. $p_k$ is the client weight, the value of which depends on the amount of data on the client and fulfills $p_k=\frac{N_k}{N}$ and $\sum_{k=1}^{K}p_k=1$.
Therefore, the training process of FL can be represented as the following steps:
\begin{itemize}
    \item \textbf{Initialization:} The server initializes the global model parameters $\theta$ and sends them to all participating clients.
    \item \textbf{Local training:} Each client $k$ trains its own model using its local data and the received global model parameters $\theta$, and computes updates $\Delta \theta_k$ to the model parameters.
    \item \textbf{Parameter aggregation:} All clients send their model parameters $\theta_k$ or gradient updates $\Delta \theta_k$ back to the central server. The server aggregates these updates based on the weights $p_k$ of each client to update the global model.
    %	\begin{equation}
        %		\label{eqb:2}
        %		\theta \leftarrow \sum_{k=1}^N p_k \theta_k
        %	\end{equation}
    %\begin{equation}
    %	\label{eqb:3}
    %	\theta \leftarrow \theta + \sum_{k=1}^N p_k \Delta \theta_k
    %\end{equation}
    \begin{equation}
        \label{eqn:cp}
        \begin{aligned}
            % &\blue{\mathrm{Model \  parameters:} \theta \leftarrow \sum_{k=1}^K p_k \theta_k,}\\
            % &\blue{\mathrm{\textbf{or} \ Model \  gradient:} \theta \leftarrow \theta + \sum_{k=1}^K p_k \Delta \theta_k.}
            \theta \leftarrow \sum_{k=1}^K p_k \theta_k, \quad
            \textbf{or}  \quad \theta \leftarrow \theta + \sum_{k=1}^K p_k \Delta \theta_k.
        \end{aligned}
    \end{equation}
    \item \textbf{Model broadcast:} The central server sends the updated global model $\theta$ to all clients.
    \item \textbf{Model iteration:} Repeat steps 2 through 4 until the stop condition is satisfied.
\end{itemize}

{Currently, FL has become increasingly widespread in real-world applications, especially in scenarios that require data privacy protection. 
\textbf{In the healthcare sector} \cite{antunes2022federated}, hospitals and medical institutions hold vast amounts of patient data, but due to privacy regulations, this data cannot be directly shared. FL allows these institutions to collaboratively train diagnostic models, such as disease prediction and medical image recognition, without exchanging sensitive data. 
\textbf{In the financial industry} \cite{nevrataki2023survey}, banks and financial institutions face a large volume of sensitive customer data (e.g., transaction records, credit information), which cannot be shared to prevent data breaches. FL enables multiple banks or financial institutions to collaboratively train models for fraud detection, credit scoring, and other applications while ensuring customer privacy is protected. 
\textbf{In the Internet of Things (IoT) domain} \cite{nguyen2021federated}, smartphones, wearable devices, and other IoT devices generate vast amounts of data. These devices are often widely distributed and face limitations in computational resources and bandwidth. FL allows these devices to process data locally and train models, reducing data transmission costs, while optimizing personalized recommendation systems or speech recognition models without uploading personal data. 
\textbf{In the field of intelligent transportation} \cite{zheng2022applications}, autonomous vehicles can share knowledge between vehicles through FL without exchanging sensitive driving data. Different vehicles can locally train models to improve algorithms for path planning, driving decision-making, and traffic prediction, thus enhancing the safety and performance of the system. Despite the widespread use of FL in various fields \cite{li2021survey}, real-world FL systems often face challenges related to heterogeneity \cite{ye2023heterogeneous}.
% Although FL has been widely used in various domains, real-world FL usually faces the problem of heterogeneity.
These include how to effectively deal with differences in data distribution across clients, how to adapt to the computational and storage capabilities of different clients while maintaining model performance, how to balance knowledge sharing with task specificity in multitasking scenarios, how to optimize communication strategies to cope with varying bandwidth and latency conditions, and how to adapt to the energy and resource constraints of different devices. }
% We next analyze the challenges posed by these heterogeneities, the resolution of which is essential to achieve efficient, scalable, and robust FL systems.

\subsection{Data Heterogeneity}
\begin{figure*}[t]
\vspace{-0.5cm}
    \centering
    \includegraphics[width=0.8\textwidth]{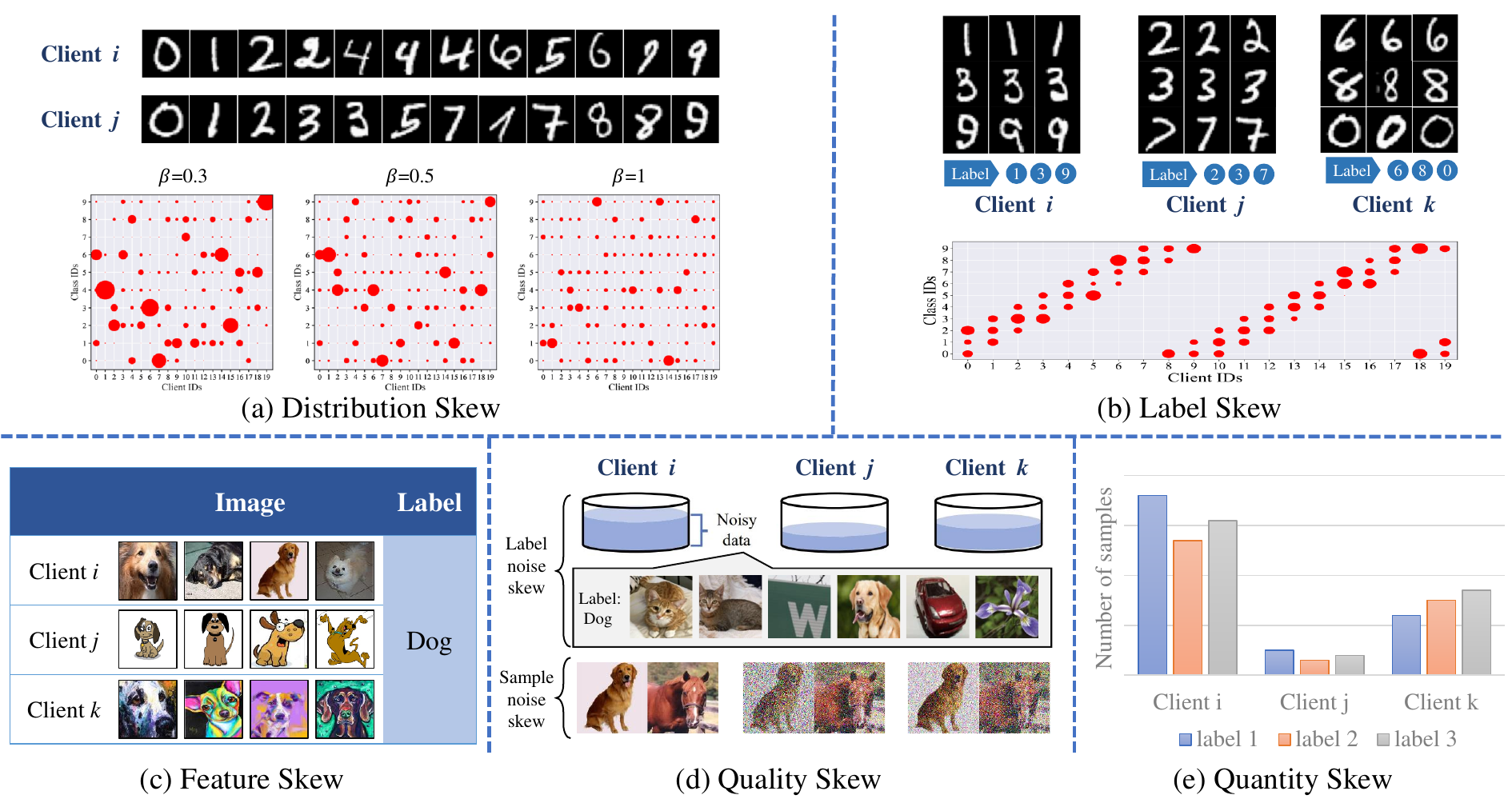}	
    \caption{Illustration of five different skew patterns in data heterogeneity.}
    \label{fig:DH}
\end{figure*}

Data heterogeneity refers to the inconsistency of data distribution among clients in FL, which does not obey the same sampling, i.e., the data are not independently and identically distributed (Non-IID) \cite{wang2020optimizing,lu2024federated}. 
To explore data heterogeneity, we categorize data heterogeneity from the perspective of client data distribution \cite{zhao2024data}. Specifically, we distinguish different categories of Non-IID data: distribution skew, label skew, feature skew, quality skew, and quantity skew, as shown in Fig. \ref{fig:DH}. 
% Numerous studies\cite{zhang2021adaptive, zhang2022fine} have found that the client's local optimization objective is inconsistent with the global optimization objective due to differences in the client's local data distribution.
Data heterogeneity may cause the local model to converge in different directions to reach the local optimum rather than the global optimum, which reduces the FL performance and leads to poor global convergence of the model \cite{gao2022feddc}.
% and these problems deteriorate the performance of the global model for each participant, and may prevent participants from engaging in FL \cite{gao2022feddc}.

\subsubsection{Distribution Skew}
In FL, data distribution skew refers to the skew in the distribution of data held by different clients, i.e., the distribution of features or joint distribution of data from different clients is inconsistent with the overall data distribution \cite{hsfed}. In this case, the data no longer obeys the assumption of independent identically distributed (IID). 
As shown in Fig. \ref{fig:DH}a, there is $P_i(x,y) \ne P_j(x,y)$ due to the difference in the data distribution of clients $i$ and $j$.
A typical example is data obeying a Dirichlet distribution \cite{li2021model} 
% \cite{wang2021fedacs}. 
% The Dirichlet distribution is a multivariate probability distribution that is often used to describe the distribution of the proportions of each category in a multicategorization problem. 
In the context of FL, the Dirichlet distribution can be used to model differences in the distribution of data labels on different clients. Where the parameter $\beta$ determines the degree of distributional difference of the data. As shown in the scatterplot in Fig. \ref{fig:DH}a, which illustrates the visualization of the difference in the distribution of data across 20 clients for different values of $\beta$.

\subsubsection{Label Skew}
Data label skew is a common Non-IID case, which refers to the skew in the distribution of data labels held by different clients. This label distribution skew means that even if the feature distributions are shared, the label distributions of different clients may be different \cite{li2021fedrs}, i.e., for clients $i$ and $j$, their label distributions $P_i(y) \ne P_j(y)$. However, for each label, the feature distributions of the data samples are similar regardless of which client they belong to, i.e., $P_i(x|y) = P_j(x|y)$. 
For example, in the data label skewing scenario shown in Fig. \ref{fig:DH}b, different users contain data from the set of handwritten digits, but each client may mainly contain a certain number of digits. The scatterplot shows a visualization of label skew for 20 clients. This data label skew can lead to a large difference in the gradient of the model uploaded back from different clients during federated training, thus affecting the convergence and generalization performance of the model.
\subsubsection{Feature Skew}
Data feature skew is the inconsistent distribution of data features across clients \cite{luo2022disentangled}, e.g., $P_i(x)\ne P_j(x)$.
In this scenario, even though the labels are consistent $P_i(y) = P_j(y)$, the distributions remain different $P_i(x|y) \ne P_j(x|y)$.
As in Fig. \ref{fig:DH}c, clients $i$, $j$, and $k$ all have the label of dog, but they are photos, cartoons, and art paintings of dogs, respectively.
This data feature skew is especially common in Vertical FL (VFL) scenarios \cite{chen2020vafl}. VFL involves multiple clients that have the same set of sample IDs, but each client holds a different set of features. In other words, each client only knows a portion of the features of the samples. 
% For example, consider two financial institutions, A and B, that collaborate to build a credit scoring model. Institution A holds the basic information and transaction history of its clients, while Institution B holds the loan history and repayment status of its clients. Although these two institutions share the same customer base, they hold different sets of data features. In this longitudinal FL scenario, data feature skew manifest itself in the form of different feature spaces held by agencies A and B.
\subsubsection{Quality Skew}
Data quality skew refers to the difference in data quality on different clients \cite{xu2022fedcorr}. This can be caused by a variety of factors as shown in Fig. \ref{fig:DH}d, such as noise in the data collection process or errors in the labelling of the collected data. In this scenario, some clients may have high-quality data that has been accurately measured and carefully labeled, while others may contain a lot of noise or inaccurate labels. High-quality data can provide more accurate information that helps the model learn effective feature representations and prediction rules. On the contrary, low-quality data may introduce misleading information that causes the model to learn incorrect feature relationships, thus reducing the model's generalization ability and prediction accuracy \cite{fang2022robust}.
\subsubsection{Quantity Skew}
Data quantity skew refers to a significant difference in the amount of data on different clients \cite{li2022federated}, as shown in Fig. \ref{fig:DH}e. This quantity skew affects the training process of a FL model because the model may tend to optimize the performance on clients with a larger amount of data while ignoring clients with a smaller amount of data.
The impact of data quantity skew on the model is mainly in two aspects \cite{qu2021experimental}: on the one hand, clients with larger data quantities may have a larger impact on the global model update, which results in the model performing better on these clients and worse on clients with smaller data quantities. On the other hand, the imbalance in the amount of data may result in the model not being able to fully learn the features of the data on all clients, thus reducing the generalization ability of the model.
% \begin{figure}[h]
%     \centering
%     \includegraphics[width=0.4\textwidth]{PPT/MH.pdf}	
%     \caption{Illustration of federated model heterogeneity.}
%     \label{fig:MH}
% \end{figure}

\subsection{Model Heterogeneity}
In the model heterogeneity scenario of FL, each client involved in learning may employ different model structures and optimizers to adapt to its own data characteristics and computational resources \cite{li2019fedmd, li2020federated}, as shown in Fig. \ref{fig:MH}. This model heterogeneity increases the complexity of FL systems, but also provides flexibility to adapt to diverse device and task requirements. We will next address model structure heterogeneity and model optimizer heterogeneity, respectively.

\subsubsection{Model Structure Heterogeneity}
Model structure heterogeneity refers to the fact that different clients use different model structures to adapt to their respective data characteristics and computational resources \cite{alam2022fedrolex}. This heterogeneity mainly stems from the diversity of devices and the diversity of task requirements in the real world. For example, in a FL system, some clients may be servers equipped with high-performance GPUs capable of handling complex deep learning models such as CNNs or transformer models for image recognition or natural language processing tasks. While other clients may be resource-constrained edge devices, such as smartphones or IoT sensors, which may only be able to run lightweight models, such as shallow neural networks or decision trees, to reduce computational and storage overhead

Moreover, even for the same type of task, different clients may choose different model structures based on the characteristics of their local data \cite{ yi2023fedgh}. For example, in the task of performing handwritten digit recognition, some clients may use a standard multilayer perceptron (MLP) model, while others may employ a convolutional neural network model with spatial feature extraction capabilities. This structural difference results in different dimensions and properties of the model parameters, making it difficult to directly apply traditional FL algorithms, such as FedAvg, which typically assume that all clients use the same model structure.

 % \blue{Therefore, typical FL methods (e.g., FedAvg) are not available when there is model structure heterogeneity among clients. To solve this problem, KD \cite{hinton2015distilling} is the key technique that utilized in most works on model heterogeneity. KD is originally proposed for transfering knowledge from a teacher model to a student model, while ensuring the performance of the student model is close to the teacher model. Therefore, KD requires the output of the student model to approach that of the teacher model on the same input. Therefore, when applying KD to handle model heterogeneity, the participants exchange the outputs rather than parameters of their private models, and learn knowledge from other participants through KD.}

% Therefore, typical FL methods (e.g., FedAvg) are not available when there is model structure heterogeneity between clients. knowledge distillation (KD) \cite{hinton2015distilling} is the key technique used in most model heterogeneity studies. KD is designed to transfer knowledge from the teacher model to the student model while ensuring that the performance of the student model is close to that of the teacher model. Thus in heterogeneous scenarios, clients exchange the outputs of their private models instead of parameters and learn knowledge from other participants through KD.

\subsubsection{Model Optimizer Heterogeneity}
Optimizer heterogeneity refers to the fact that different clients use different optimization algorithms or parameters to train the model \cite{castiglia2023flexible, abdelmoniem2023comprehensive}. Due to the different computational power and memory constraints of different devices, some clients may use optimizers with smaller memory footprints and computational costs, such as Stochastic Gradient Descent (SGD), while others may use more sophisticated optimizers, such as Adam or RMSprop, to achieve faster convergence. Moreover, even with the same optimizer, different clients may employ different learning rates or other hyperparameter settings to suit their local data distribution and training conditions. This optimizer heterogeneity increases the complexity of FL, as coordination between different optimization strategies is required to ensure effective global model updating and convergence.

\subsection{Task Heterogeneity}
\begin{figure*}[h]
    \centering
    \vspace{-0.5cm}
    \includegraphics[width=0.9\textwidth]{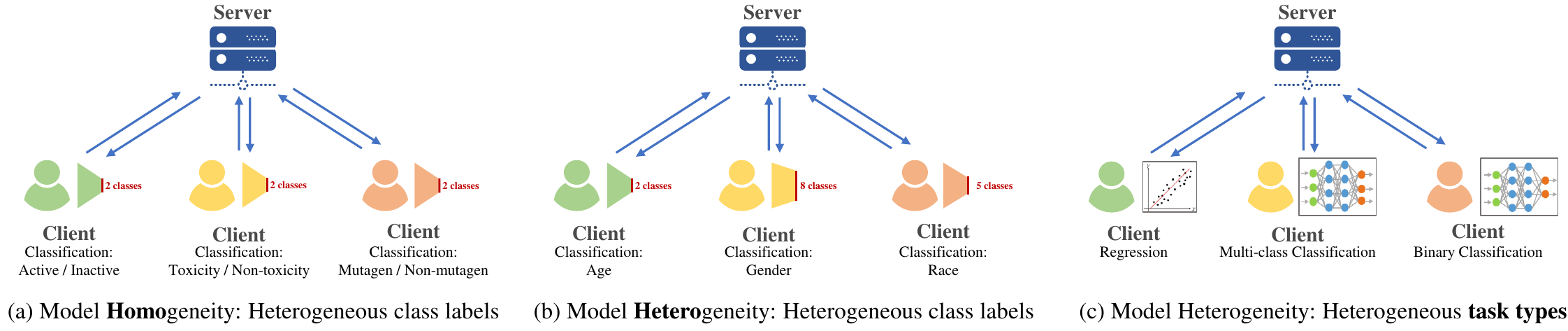}	
    \caption{FL task heterogeneity.}
    \label{fig:TH}
\end{figure*}

In FL, task heterogeneity refers to the fact that clients involved in the learning process may face different task requirements or objective functions \cite{smith2017federated}. This heterogeneity may stem from the environment in which the clients are located, differences in user behavior, or specific application scenario requirements, resulting in different objective functions or learning tasks that need to be optimized for each client.
%For example, in a FL application in the healthcare domain, different hospitals may focus on different types of disease diagnosis, and thus their learning tasks and objective functions may be different. As shown in Fig. \ref{fig:TH}, Fig. \ref{fig:TH}a demonstrates that when clients have similar datasets such as face dataset, individual clients have heterogeneous classification labels. Fig. \ref{fig:TH}b task types are heterogeneous, including classification and regression tasks.
For example \cite{yao2022benchmark}, in a FL application in the medical domain, different hospitals may focus on different types of disease diagnosis, so their learning tasks and objective functions may be different, e.g., Fig. \ref{fig:TH}a demonstrates that clients have homogeneous models but the distributions have labels for different medicines. Fig. \ref{fig:TH}b demonstrates when clients have similar datasets such as face dataset, and individual clients have heterogeneous classification labels. Fig. \ref{fig:TH}c Task types are heterogeneous, including classification and regression tasks.

Task heterogeneity introduces additional complexity to FL because traditional FL frameworks typically assume that all devices jointly optimize a global objective function \cite{sarcheshmehpour2023networked}. 
Task heterogeneity requires FL systems to be able to efficiently coordinate the learning process across devices to ensure a balance between the generalisation capabilities of the global model and the personalised needs of the client. Such coordination may require appropriate adaptation of the model updating and parameter aggregation processes to suit the characteristics and needs of different tasks \cite{liao2024swiss}.
% To cope with task heterogeneity, FL systems need to employ more flexible algorithms and policies that allow different devices to personalize model training based on their own task requirements.
% In addition, task heterogeneity also requires that FL systems are able to effectively coordinate the learning processes of different devices to ensure a balance between the generalization capabilities of the global model and the individual needs of each client. Such coordination may require appropriate adaptation of the model updating and parameter aggregation processes to suit the characteristics and needs of different tasks.

\subsection{Communication heterogeneity}
\begin{figure*}[htbp]
	\centering
  \vspace{-0.5cm}
	\begin{minipage}{0.333\linewidth}
		\centering
		\includegraphics[width=0.9\linewidth]{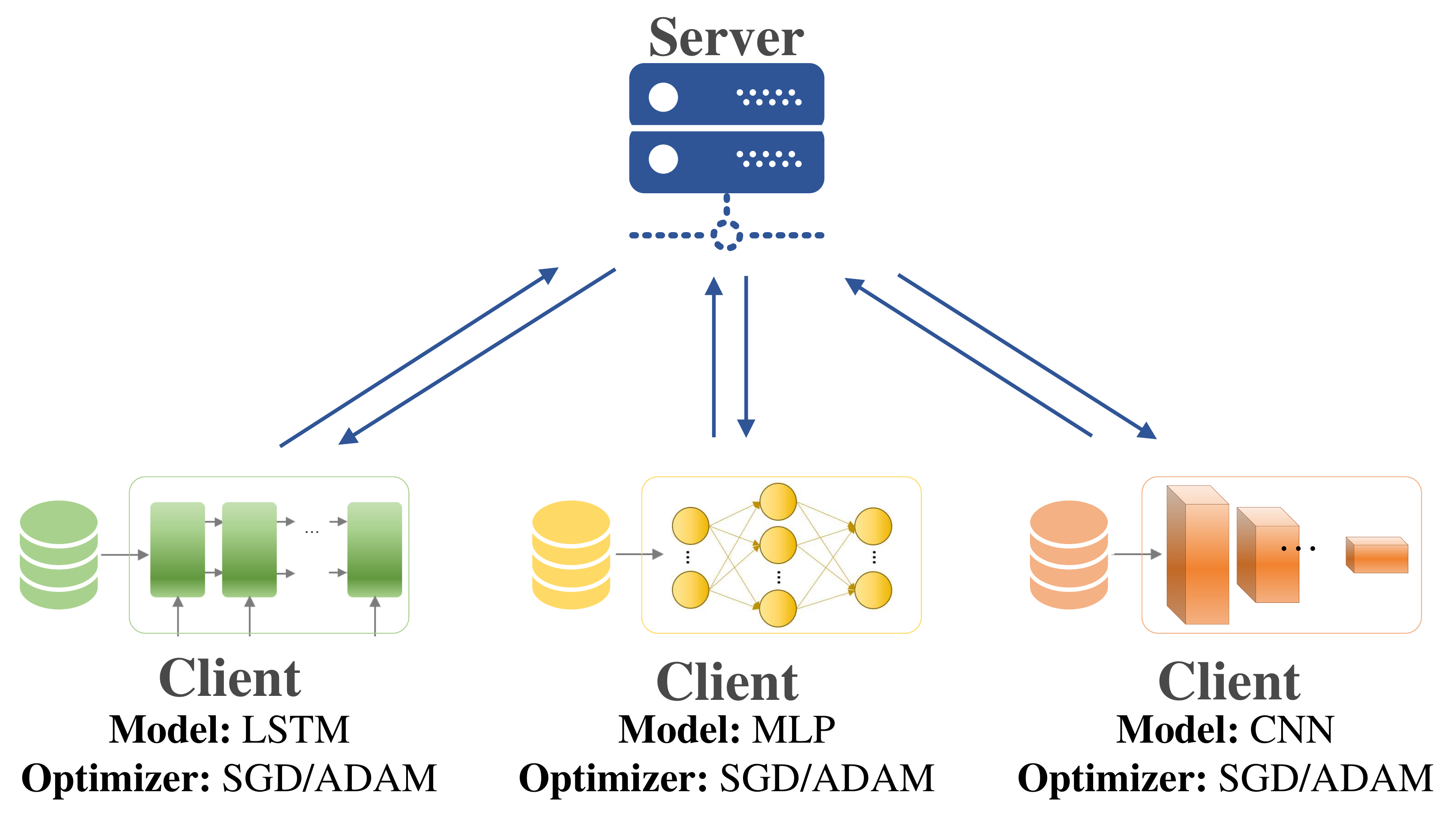}	
		\caption{FL model heterogeneity.}
		\label{fig:MH}%文中引用该图片代号
	\end{minipage}
	\hspace{-0.5cm}
	\begin{minipage}{0.333\linewidth}
		\centering
		\includegraphics[width=0.9\linewidth]{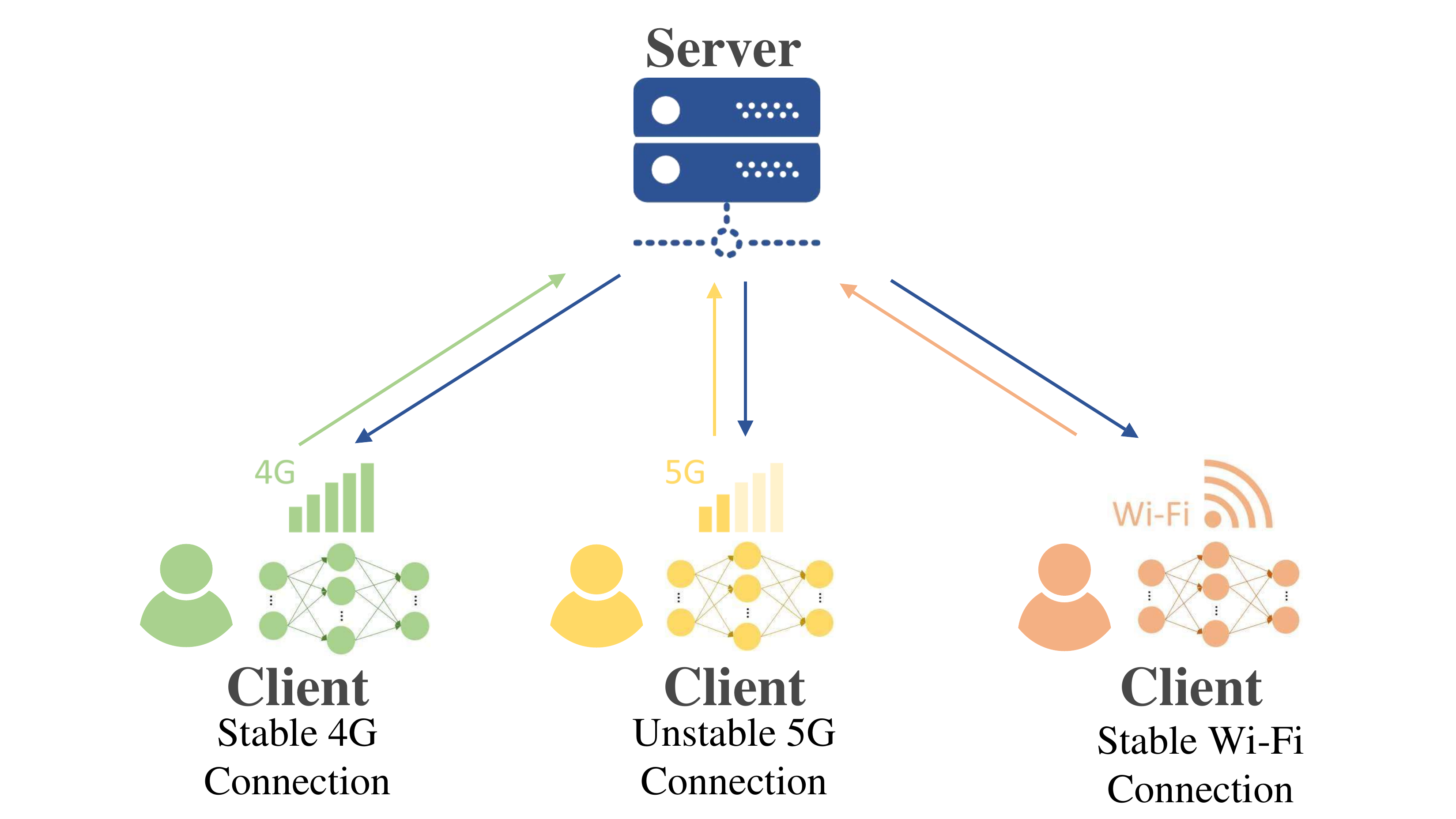}	
		\caption{FL communication heterogeneity.}
		\label{fig:CH}%文中引用该图片代号
	\end{minipage}
 	\hspace{-0.5cm}
	\begin{minipage}{0.333\linewidth}
		\centering
		\includegraphics[width=0.9\linewidth]{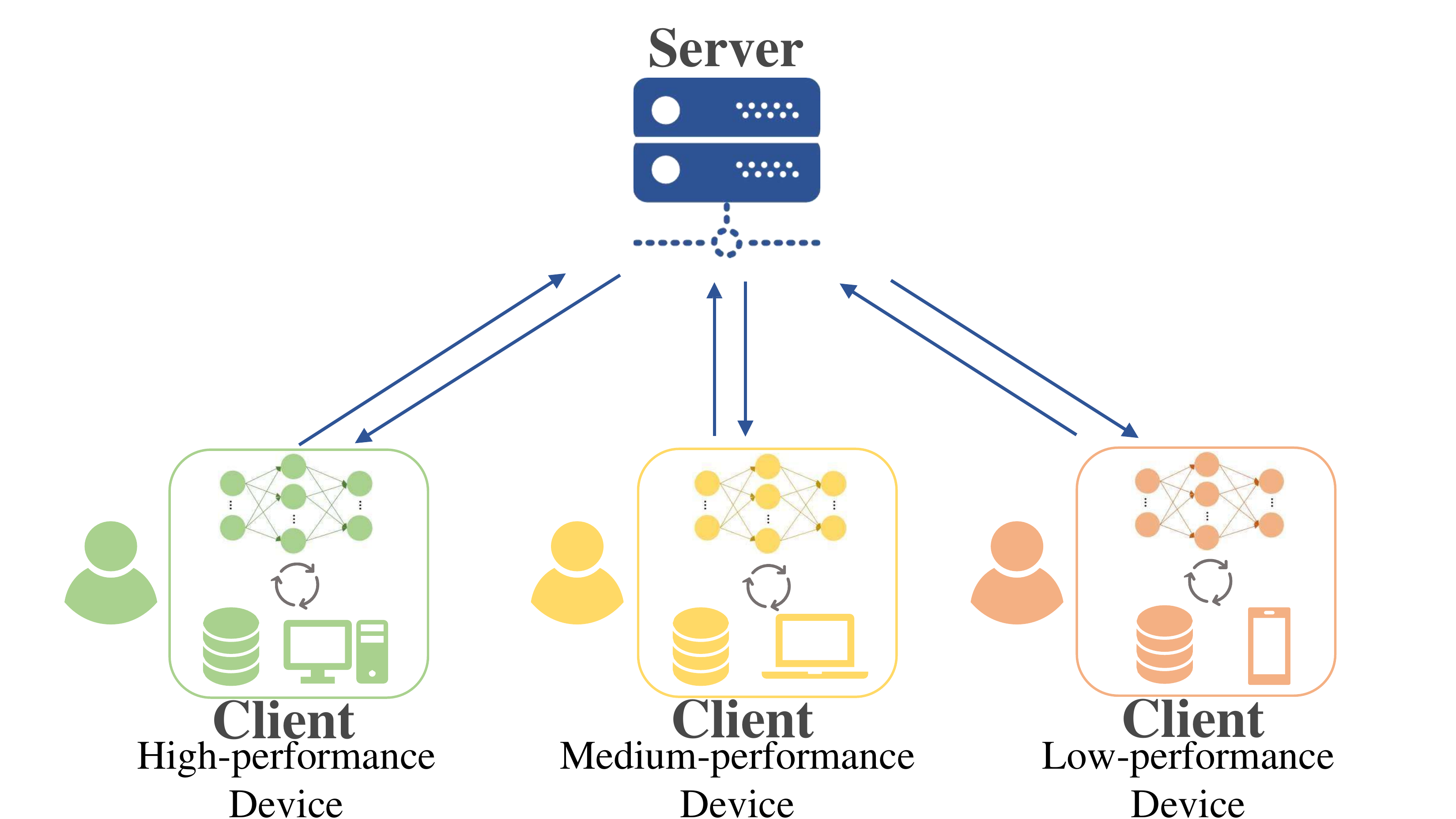}	
		\caption{FL device heterogeneity.}
		\label{fig:DevH}%文中引用该图片代号
	\end{minipage}
\end{figure*}

Communication heterogeneity refers to the differences in communication capabilities of the devices involved in the learning process \cite{abad2020hierarchical}. Such differences are mainly in terms of network bandwidth, latency and connection stability, which directly affect the transmission efficiency of model parameters and the synchronization of the whole learning process. For example, devices are usually deployed in different network environments with different network connection settings (3G, 4G, 5G, Wi-Fi) \cite{diao2020heterofl, abdellatif2022communication}, as shown in Fig. \ref{fig:CH}. Devices with high bandwidth and low latency can quickly upload and download model parameters, thus completing a round of model updates in a short period of time. On the contrary, devices with limited bandwidth and high latency may face longer waiting time when transmitting model parameters, resulting in the training progress of the whole system being affected. The heterogeneity of communication increases the communication cost and complexity to some extent \cite{nishio2019client}. In addition, differences in connection stability can also have an impact on the efficiency of FL. Unstable connections may result in lost or delayed model parameter updates, which in turn affects the convergence speed and accuracy of the model. For example, in Fig. \ref{fig:CH}, though the middle client accesses the network through 5G connection, the connection is unstable and has a higher risk of packet loss, which can hinder the process of FL.
% \begin{figure}[h]
%     \centering
%     \includegraphics[width=0.45\textwidth]{PPT/CH.pdf}	
%     \caption{\blue{Illustration of federated communication heterogeneity.}}
%     \label{fig:CH}
% \end{figure}

\subsection{Device heterogeneity}
In the application scenario of FL, device heterogeneity is a crucial issue that cannot be neglected \cite{wang2021device}. Due to the wide variety of client devices participating in FL, there are significant differences in computing power, storage resources, energy supply, etc., which leads to device heterogeneity \cite{pfeiffer2023federated}. As shown in Fig. \ref{fig:DevH}, each client is equipped with a different type of device, resulting in a significant performance span between the clients. This heterogeneity leads to inconsistent efficiency and capability of different devices in processing data and training models, which may affect the performance and efficiency of the whole FL system. For example, high-performance devices such as servers or high-end smartphones are able to process complex models and large data volumes quickly due to their powerful processors and ample storage space, thus taking on more computational burden during FL and speeding up the entire learning process \cite{xu2021optimizing}. On the contrary, low-performance devices, such as low-end smartphones, are less capable of model training and data processing due to the limitations of computational resources and storage space, and may also slow down the progress of the entire FL system, which may become the bottleneck of system efficiency. This is because the system usually needs to wait for all devices to complete their local updates before global model aggregation can take place. In addition, devices with weaker computational power may face greater pressure on energy consumption when processing complex models, which is especially critical in energy-constrained environments.

%     \begin{figure}[h]
%     \centering
%     \includegraphics[width=0.48\textwidth]{PPT/DevH.pdf}	
%     \caption{\blue{Illustration of federated device heterogeneity.}}
%     \label{fig:DevH}
% \end{figure}

% Therefore, FL algorithms and system design need to take into account the heterogeneity of computational capabilities among devices, and ensure efficient operation and fairness of the system through strategies such as reasonable task allocation, model simplification, and resource optimization to ensure that all devices can effectively participate in the learning process, while maintaining the coordination and fairness of the overall system.
\section{Advanced Methods Discussion}

\subsection{Data-Level Methods}
Doing operations at the data level is often seen as an important means to address data heterogeneity. In this section, we categorize the data-level approaches into three categories: data cleansing, data augmentation, and representation learning, and present their representative approaches respectively.
\subsubsection{Data Cleaning}
Data cleaning in FL can help improve the training effect and final performance of the model. Data cleaning mainly refers to correcting errors and inconsistencies in datasets, including dealing with issues such as missing values, outliers, format errors, and duplicate data. 
%In FL systems, the selection of training samples has a significant impact on model performance, for example, selecting participants whose dataset has erroneous samples, skewed classification distributions, and low content diversity will result in low accuracy and model instability.

Li \textit{et al.} \cite{li2021sample} presents an optimization problem i.e., selecting a high-quality set of training samples for a FL task under a given budget in a privacy-preserving manner without knowledge of the participants' local data and training process. 
% Ma \textit{et al.} \cite{ma2020federated} proposes a federated data cleaning protocol, called FedClean, for edge intelligence scenarios that aims to achieve data cleaning without compromising data privacy. 
% FedClean Different edge nodes first generate a Boolean share of their data and distribute it to two clean servers. These two servers then run the FedClean protocol to compute and sort the data entry scores.
Faced with the problem of noisy data labels, Chen \textit{et al.} \cite{chen2022robust} proposes a robust FL by considering two networks with different data distributions. The core idea is that the edge devices learn to assign local weights of the loss function in the noisy labeled dataset and cooperate with the central server to update the global weights as a way to help edge nodes that do not have a clean dataset to reweight their noisy loss functions. 
Similarly, Duan \textit{et al.} \textit{et al.} \cite{duan2022fed} proposes a new data filtering approach to deal with label noise in joint learning, called FED-DR-Filter, which focuses on identifying clean data by exploiting the correlation of global data representations. 
Tsouvalas \textit{et al.} \cite{tsouvalas2024labeling} proposes FedLN framework enough to deal with labeling noise in different FL training phases, i.e., FL initialization, on-device model training, and server model aggregation, which is able to adapt to the different computational power of the devices in the FL system. 
{Li \textit{et al.} \cite{li2024feddiv} proposed FedDiv to address the challenge of FL with noisy labels, and they designed a global noise filter with a joint noise filter for efficiently identifying samples with noisy labels on each client, which improves the stability of the local training process.} 
Xu \textit{et al.} \cite{xu2022safe} proposed a collaborative data filtering method Safe to deal with poisoning attacks, using the alternating direction method of multipliers to detect the attacked devices whose training process will be interrupted, and using the K-mean clustering algorithm to cluster the trusted data to find out and filter out the poisoned data from the attacked devices. 

\subsubsection{Data Augmentation}
Data augmentation plays a key role in FL, especially when dealing with Non-IID data distributed across multiple devices. Data augmentation not only increases the amount of data, but also increases the diversity of the data.

Hao \textit{et al.} \cite{hao2021towards} proposes a FL system and derives 2 variants that can use zero-sample data augmentation on underpresented data to mitigate data heterogeneity. 
Chen \textit{et al.} \cite{chen2023fraug} proposes FRAug, a federated representation augmentation method, to address the data heterogeneity challenge. FRAug optimizes a shared embedding generator to obtain client consensus, and then locally transforms the output synthetic embeddings to client-specific embeddings in order to increase the training space for each client. 
Lewy \textit{et al.} \cite{lewy2022statmix} proposed StatMix, an enhancement method using image statistics to improve the results of FL scenes. 
{Xin \textit{et al.} \cite{xin2020private} proposes FL-GAN, a generative adversarial network model for differential privacy based on joint learning. by combining the Lipschitz limit with differential privacy sensitivity, the model can generate high-quality synthetic data without sacrificing the privacy of the training data.} 
Zhang \textit{et al.} \cite{zhang2024federated} proposes a distributed data augmentation method that combines generative adversarial networks and federated analysis to enhance the generalization ability of trained FDG models, called FA-FDG. 
Generating more overlapping data by learning the features of limited overlapping data and many locally existing non-overlapping data, Zhang \textit{et al.} \cite{zhang2022data} proposed a data augmentation method FedDA based on generative adversarial networks to increase the amount of training data.
Similarly, Xiao \textit{et al.} \cite{xiao2024distributed} proposes FeCGAN, a distributed generative adversarial network for multiple participants with insufficient data overlap.
\subsubsection{Representation Learning}
Representation learning aims to learn useful data representations (or features) from dispersed data sources to help improve the predictive performance and generalization of models.

To address the problem of data heterogeneity, Liu \textit{et al.} \cite{liu2020secure} introduces federated transfer learning, which does not compromise user privacy and can transfer knowledge across domains in a data federation. This enables the target domain to utilize the rich labels of the source domain to build flexible and efficient models.
{Li \textit{et al.} \cite{li2021fedmask} proposes FedMask, a FL framework for efficient communication and computation, where each device can learn an efficient personalized structured sparse model.} 
Zhang \textit{et al.} \cite{zhang2023federated} proposes for Federated Unsupervised Representation Learning (FURL) to learn generic representation models without supervision while preserving data privacy. FURL solves the problem of inconsistent and misaligned representation space through a dictionary module and an alignment module. 
Similarly, Yan \textit{et al.} \cite{van2020towards} uses unsupervised representation to pre-train deep neural networks in a federated environment using unlabeled data, which allows for the extraction of discriminative features using pre-trained networks. 
{Wu \textit{et al.} \cite{wu2022practical} proposes a communication-efficient FL algorithm, FedOnce, which is based on the privacy-preserving technique of momentary accounting and employs an unsupervised learning representation that enables better results to be achieved with only one-time communication between the parties.}
{Li \textit{et al.} \cite{ li2021model} proposed MOON to correct for local training of individuals using similarities between model representations, and thus reduce the impact of data heterogeneity.}
Jang \textit{et al.} \cite{jang2022fedclassavg} proposes a personalized FL approach that applies local feature representation learning to stabilize the decision boundary and improve the local feature extraction capability of the client. 
{Tan \textit{et al.} \cite{tan2022fedproto} proposes FedProto, which only passes the prototype to the client without passing model parameters and gradients, which can be available when it is data heterogeneity and model heterogeneity.}

% \subsubsection{Knowledge Distillation}
\subsection{Model-Level Methods}
Model-level methods means improving FL by adopting some techniques that only apply to model parameters, model outputs, etc., and make minor changes to the input data and FL architecture.  Specifically, we classify these methods based on their techniques, including parameter decoupling, model compression, knowledge distillation and federated optimization.

\subsubsection{Parameter Decoupling}
Parameter Decoupling based methods regard the whole model parameters as the composition of components, which play different roles in the process of training or inferring. Depending on the goal of parameter decoupling, different methods usually have different granularities. Some methods only regard the whole model as the composition of a backbone model and a head model, while some others decouple the model in a layer-wise or channel-wise, even coordinate-wise manner. Arivazhagan \textit{et al.} {\cite{arivazhagan2019federated} propose FedPer, which split global model into base layers and personalized layers. Base layers are layers closer to the input side while the personalized layers are the remaining layers. Personalized layers is customized while base layers are shared and transmitted. Therefore, FedPer can also to some extent handles task heterogeneity and communication heterogeneity. }Based on the belief that data from all parties involved in FL often share global feature representations while reflect distribution and label skew in the labels, Collins \textit{et al.} \cite{collins2021exploiting} propose FedRep, which separate model into representation part and head part. The clients train a global representation in a federated manner and then train their own head based on the learned representation. {Luo \textit{et al.} \cite{luo2021no} empirically verify that distribution and label skew impact more on deeper layers, CCVR, which utilizes regularization and calibration in local training to reduce weight bias, and trains a GMM to generate virtual representations for further classifier training, which can help reducing the impact of data heterogeneity.} To solve the problem of long-tailed data in FL, 
% Shang \textit{et al.} \cite{shang2022federated} propose CReFF, where clients upload real feature gradients, and then the server use them to optimize federated feature gradients for re-training classifier. Comparing to above works that enable shared global representations of features, 
Liang \textit{et al.} \cite{liang2020think} propose LG-FedAvg, where clients share a common architecture of classifier, and each client learns compressed local representation for training a classifier suitable for local representations of all clients. Learning such local representations on the clients can also help alleviating task heterogeneity, encouraging the clients to learn and extract representations suitable for global classifiers. Xu \textit{et al.} \cite{xu2023personalized} propose FedPAC, which encourage each client to learn global task variant information by regularizing local representations to approach the global feature centroid. Besides, it also trains personalized classifiers and collaborates similar clients, reducing inter-client feature variance. {In order to train personalized client models under feature skew, distribution skew and other scenarios, 
%comparing to previous layer-wised personalization methods \cite{arivazhagan2019federated} \cite{collins2021exploiting} \cite{liang2020think} that rely on prior knowledge, 
Shen \textit{et al.} \cite{shen2022cd2} propose CD$^2$-pFed, which implement channel-wised personalization by channel-decoupling, and utilize personalized-weights to dynamically control the degree of personalization in each layer. Besides, CD$^2$-pFed also utilizes cyclic-distillation to mutually distill local and global model representations, bridging the semantic gap between the decoupled channels.}
%\cite{ma2022layer} maintains a hypernetwork on the server to generate aggregation weights for each layer based on the importance of each layer between different clients, thereby implementing layer-wised personalization for the global model. 
% Recently, Chen \textit{et al.} \cite{chen2024watch} propose APH method, where clients can get multiple initialized parameters from existing projection heads, and then fine-tune and average them, thus eliminating the bias caused by data heterogeneity. 

Some works based on parameters decoupling are also proposed for handling model heterogeneity. To reduce the memory cost of clients, Pillutla \textit{et al.} \cite{pillutla2022federated} propose to divide the model parameters into two parts: shared parts and personal parts. Therefore, they propose FedSim and FedAlt, representing training both part of parameters in a simultaneous manner or an alternate manner, respectively. {Yi \textit{et al.} \cite{yi2023fedgh} propose model heterogeneous method FedGH, where clients share a generalized head. The clients upload local average representation, which is utilized for individually training of the generalized head on the server side.} However, the above works require some specific parts of the model structure to be identical among the clients, which can't be used for the scenarios with thorough model heterogeneity. Recently, to achieve personalization when models are heterogeneous, Wang \textit{et al.} \cite{wang2024towards} propose pFedHR. The server decomposes and then groups the client's models layer by layer in each round. Afterwards, the server reassembles the candidate models using the grouped layers and fine-tunes them using a public dataset.

{Parameter decoupling can also be utilized for handling device heterogeneity, which is performed by reducing the size of the model. To alleviate the impact of device heterogeneity and communication heterogeneity, Diao \textit{et al.} \cite{diao2020heterofl} propose HeteroFL, where the server determines the complexity level of each client based on its computation and communication capabilities, and assigns sub-models of different sizes to clients according to their complexity level.} However, comparing to traditional FL methods, HeteroFL suffers from significant performance degradation. Recently, Yi \textit{et al.} \cite{yi2024fedp3} point out that even pruned models can also reveal the real architecture of the global model. Therefore, they propose a novel framework FedP3, where clients only upload selected model parameters back to the server for privacy protection. Pfeiffer \textit{et al.} \cite{pfeiffer2024aggregating} empirically observe that the techniques based on sub-models suffer from performance degradation because the co-adaptation between parameters is broken. Thus, they propose SLT. All clients can train the same parameters, but they begin at early layers with a down-scale head. After early layers are well trained, then SLT squeeze the parameters of early layers and expand the head layers for further training.

\subsubsection{Model Compression}
%	Model compression means compressing the size of the model. 
Model compression methods are aiming at compressing the size of a model and is commonly used to solve model, device and communication heterogeneity issues, including sparsification, quantization and factorization.
\textbf{Sparsification} usually means identify and retain the meaningful enough parameters, while setting other parameters to some constants. In this way, the parameter matrices are transformed to be sparse, and only those meaningful enough parameters are transmitted. Different from sparsification, \textbf{quantization} keeps the parameters dense and focuses on reducing the bit-width occupation of each parameter. A representative quantization method is SignSGD \cite{bernstein2018signsgd}, which only retains the sign bit of each parameter. \textbf{Factorization} decompose a large matrix into the product of multiple low-rank matrices or vectors, and thus compress the size of the whole model with limited precision degradation. {To compress both upstream and downstream communications, Sattler \textit{et al.} \cite{sattler2019robust} point out that sparsification is more robust to Non-IID data and thus propose Sparse Ternary Compression (STC), which further quantizes the remaining top-$k$ elements to the mean population magnitude and only transfer the remaining elements.} Similar to the goal of \cite{sattler2019robust}, Amiri \textit{et al.} \cite{amiri2020federated} propose Lossy FL (LFL). The server broadcast the quantized version the global model parameters to the clients, and the uploading parameters are also quantized. %\cite{malekijoo2021fedzip} proposes FedZIP, which process the uploading parameters with sparsification based on Top-z pruning, then quantization based on k-means clustering on parameters, and finally encoding with Huffman Encoding and other encoding methods. But it only considers upstream communications. 
%\cite{chen2021joint} combines probabilistic device selection with quantization and wireless resource allocation. It assumes orthogonal frequency-division multiple access (OFDMA) technique is used for the devices to send their local FL parameters, and each device can only be allocated with one resource block, which is determined by the server, the same as device selection. For compression the model, it adopts dithered vector quantization based on random lattices. 
{Shah \textit{et al.} \cite{shah2021model} propose to compress both downstream and upstream communications, but it goes further. To recover the loss of sparsity after aggregation, each client can perform sparse reconstruction for the averaged model, and it can also directly begin local training with the received sparse model.} 
%Moreover, the clients with sufficient computation resources can train a binary mask for the model based on two-time-scale theory, and the clients with less sufficient computation resources can firstly initialize client model with a sparse subnetwork via a saliency criterion and then only train the non-zero parameters. 
Recently, Tang \textit{et al.} \cite{tang2024z} point out that SignSGD is vulnerable to Non-IID data, and current corresponding methods cannot support multiple local SGD updates like FedAvg. Thus, they propose to inject some random noise before the sign operation to control the bias brought by it, and further propose the first sign-based FedAvg algorithm $z$-SignFedAvg. 
% \cite{huang2023stochastic} reformulate SCAFFOLD and propose SCALLION and SCAFCOM, which compress the uploading information with unbiased and biased compressor, respectively. 

{To combat the challenge of model and device heterogeneity, Yao \textit{et al.} \cite{yao2021fedhm} propose FedHM, where the server compresses the large global model via low-rank factorization for each client to satisfy its resource availability. The compressed model can be transformed back to full-rank model for aggregation. Jeong \textit{et al.} \cite{jeong2022factorized} propose Factorized-FL, which factorize model parameters into basis vectors. It can make aggregation operation more robust to different labels and data, helping address the issue of data heterogeneity. Recently, Chen \textit{et al.} \cite{chen2024fed} propose a personalized FL framework Fed-QSSL, where clients can perform low-bitwidth quantization training locally to satisfy its device constraint, and tackle data heterogeneity with distributed self-supervised learning.}

\subsubsection{Knowledge Distillation}
Knowledge distillation \cite{hinton2015distilling} is originally proposed for transferring knowledge from a larger model to a smaller model, and thus reducing the number of parameters. Generally speaking, This process requires the output of the smaller model to approach the larger one. Therefore, KD can also be used for knowledge transferring between participants in FL.

Knowledge distillation is mostly adopted for handling model heterogeneity. {Li \textit{et al.} \cite{li2019fedmd} firstly consider model heterogeneity in FL and propose FedMD. FedMD introduces a public dataset, on which the clients upload their private models' soft labels to the server for aggregation as consensus. The clients utilize the consensus as the output of the teacher model and transfer it to their own private model by KD on public dataset, and then perform a few epochs of training on their own private dataset. Therefore, the clients can safely share their knowledge without exposing model structure and transmitting parameters.} Cronus proposed by chang \textit{et al.} \cite{chang2019cronus} is similar to FedMD, but it takes into account the possibility of attacks during aggregating the knowledge and improves the process of aggregation to be robust. Li \textit{et al.} \cite{li2021fedh2l} propose FedH2L, where the clients can learn knowledge from others by accessing their public datasets and perform KD, and correct the gradients to reduce the conflicts of gradients between private and public dataset caused by data heterogeneity. Cheng \textit{et al.} \cite{cheng2021fedgems} consider how to train a larger global model in the case of heterogeneous client models and propose FedGEMS, which selective upload client outputs on public dataset, and establish a selective knowledge fusion mechanism on the server for effective knowledge transfer. {To alleviate the impact of feature skew and model heterogeneity, Huang \textit{et al.} \cite{huang2022learn} propose FCCL, where each client constructs a cross-correlation matrix according to other clients’ soft labels on public dataset, and utilize dual-domain KD to alleviate catastrophic forgetting caused by domain shift. For scenarios with model heterogeneity and quality skew, Fang \textit{et al.} \cite{fang2022robust} proposes RHFL, where clients combine cross entropy and reverse cross entropy to alleviate the impact of label noise, and perform KD on a public dataset to reduce the knowledge gap between clients.}

The above works all introduce public datasets, but in reality, such datasets are not available in most cases. Therefore, some other works mainly focus on utilizing KD without relying on public datasets. {Shen \textit{et al.} \cite{shen2020federated} propose FML, which shares a small meme model between the server and all clients. Each client performs mutual learning between its private model and the meme model, and then upload the parameters of the meme model for aggregation. Therefore, knowledge can be shared between the clients through the meme model. Wu \textit{et al.} \cite{wu2022communication} propose FedKD, which also utilizes a global model for KD on logit and representation space, and the communication cost of global model is reduced by performing SVD-decomposition on its parameters.} He \textit{et al.} \cite{he2020group} propose FedGKT for training large CNNs when the performance of edge devices is limited. The clients upload computed features and soft labels as well as real labels to the server. The server utilize them to conduct KD, and send its soft labels over the features back to the clients for local KD. FedDKC proposed by Wu \textit{et al.} \cite{wu2024exploring} is similar to FedGKT, but they propose two mechanisms for client knowledge congruence to mitigate the impact of knowledge differences between clients. Zhang \textit{et al.} \cite{zhang2023towards} propose FedGD, which introduces a distributed trained generator fixed on the server to generate intermediate features. The clients upload classifier parameters and soft labels for generator training, and perform KD on the generated intermediate features. Zhang \textit{et al.} \cite{zhang2022feddtg} also propose a generator-based method called FedDTG. Unlike FedGD, in FedDTG, the generator directly generates samples in raw data space and is not fixed on the server, but is trained by adversarial training with local classifiers and discriminators on the client side.

Apart from model heterogeneity, knowledge distillation can also be utilized for handling other types of heterogeneity. To alleviate the impact of data heterogeneity, Yao \textit{et al.} \cite{yao2021local} propose FedGKD, where the server stores global model parameters aggregated in recent multiple historical rounds. During local training, the knowledge from the historical global models is transferred to the client models through KD. {Zhang \textit{et al.} \cite{zhang2022fine} propose FedFTG, where the server maintains a generator trained with clieng models for generating pseudo data, and the server can perform KD based on the pseudo data to transfer knowledge from the client models to the aggregated global model.} Zhu \textit{et al.} \cite{zhu2021data} propose FedGEN, where the server trains a light-weighted generator for representation generation. While local training, the clients utilize the generated representations for KD. Recently, {to combat severe label noise caused by highly contaminated clients, Lu \textit{et al.} \cite{lu2024federated} propose FedNed, which identifies noisy clients and requires them to train models with noisy labels and pseudo-labels simultaneously. Then the models trained with noisy labels act as bad teachers for negative distillation, while the ones trained with pseudo-labels are used for aggregation. To learn personalized models under data heterogeneity, Chen \textit{et al.} \cite{chen2024spectral} propose a method to capture similarity between generic model and personalized models by spectral distillation. Moreover, it establishes a co-distillation framework to bridge the knowledge of the generic model and personalized models.} 
%In order to tackle label distribution skew and feature skew simultaneously, \cite{cai2024fed} propose Fed-CO$_2$, which establishes an online model for domain-invariant knowledges, and personalized offline models for domain-specific knowledges. Moreover, Fed-CO$_2$ also introduce inter-client knowledge transfer mechanism to facilitate mutual learning between the online and offline models, and intra-client knowledge transfer mechanism to help improve the generalization ability of the online model.

% For efficient communication, \cite{sattler2020communication} proposes CFD based on federated distillation. CFD introduces distill data curation, soft-label quantization and delta-decoding to compress knowledge uploaded by clients, and it also quantify the average soft label sent from the server to the clients to reduce downstream communication cost. 

\subsubsection{Federated Optimization}
Some methods introduce other techniques to the process of federated optimization, which can help the models to more adaptable to the local distributions or the global distribution. We mainly discuss three federated optimization techniques applies in model-level approaches: Multi-task learning, meta-learning and regularizing.

\textbf{Federated multi-task learning}. The core idea of multi-task learning is to train a model simultaneously to perform multiple related tasks, thereby improving the model's generalization ability. By regarding each client's local training as a separate task, multi-task learning techniques can help clients to learn personalized model with high generalization ability. Smith {\textit{et al.} \cite{smith2017federated} extend MTL method CoCoA from traditional distributed learning to federated environment, and propose MOCHA, which considers each client’s local training as a task. MOCHA learns a separate model for each client instead of a global model, which is robust to data heterogeneity.} Corinzia \textit{et al.} \cite{corinzia2019variational} propose to construct a Bayesian network between the server and the clients, and utilize variational inference for its optimization. Comparing to MOCHA, it requires no additional regularization term in local training objective and thus ensuring network-agnostic FL. Besides, it also extends FMTL to scenarios with non-convex objectives. To learn a personalized model for devices that later join FL, Li \textit{et al.} \cite{li2019online} propose OFMTL, which helps the new devices to learn with the assistance of weight matrix and precision matrix constructed by old devices, requiring no presence of earlier devices. Dinh \textit{et al.} \cite{dinh2021fedu} propose FedU, which introduce Laplacian regularization to determine the correlation between clients in FMTL. Besides, They also propose a decentralized version dFedU. Marfoq \textit{et al.} \cite{marfoq2021federated} consider the data distribution of each client as a mixture of several latent distributions. Therefore, they propose FedEM, where clients can collaboratively train several generic component models, and each client only needs to learn how to mix these components to build a personalized model. 

\textbf{Meta-learning based FL}. Meta-learning, which is also known as "learning to learn", is usually used to design and train algorithms to quickly adapt to new tasks, typically when it is very little data. The core goal of meta-learning is to improve the flexibility and adaptability of learning algorithms, enabling them to learn from previous experiences and better handle new and unknown tasks. {Chen \textit{et al.} \cite{chen2018federated} combine FL with meta-learning and then propose FedMeta, where all clients collaboratively train a initialization model. Each client upload its test loss on its query set which is used to update the initialization model. Each client can learn a personalized model from the initialization model by just a few local training, thus FedMeta can also reduce the training burden of the clients and reduce the performance degradation of the final personalized models due to the scenarios with data heterogeneity.} Recently, Jeon \textit{et al.} \cite{jeon2024federated} propose a Bayesian meta-learning based approach called MetaVD, where the server  establishes a shared hypernetwork for predicting client-specific dropout rates, which are utilized for few-shot local adaptation on clients. Leo \textit{et al.} \cite{lee2024fedl2p} propose FedL2P, which introduces meta-nets to inductively map each client’s Batch Normalization statistics to client-specific personalization hyperparameters. Scott \textit{et al.} \cite{scott2023pefll} propose PeFLL. The server is equipped with a hypernetwork, which takes the descriptor of each client as input and personalized model parameters as output. The descriptor of each client is obtained with an embedding network, taking some data points as input. Thereby PeFLL overcomes the problems of high latency and client-side computation cost. However, when the required model's size is significantly large, such hypernetwork could be hard to establish because there too many model parameters to output.

\textbf{Regularization based FL}. Regularization usually means adding a regularization term in the local training objectives and thus achieving some specific goals, such as tackling data heterogeneity. {Li \textit{et al.} \cite{li2020federated} propose FedProx, which adds a proximal term in local training objective to reduce the distance between the local model and the global model, and thus handles data heterogeneity. Besides, They also propose in-exact solution, enabling the clients to upload incompletely optimized parameters and thus alleviate the problem of device heterogeneity.} Kim \textit{et al.} \cite{kim2022multi} propose FedMLB, which separate models into several blocks, and each client form a main pathway and several hybrid pathways with the blocks. Regularization terms based on KD are formed with the output of the main pathway and the hybrid pathways. {By performing KD on hybrid pathways between the global model and the local model, FedMLB can bridge the knowledge gap caused by data heterogeneity, especially distribution skew and label skew.} Shoham \textit{et al.} \cite{shoham2019overcoming} point out that due to data heterogeneity, the problem of forgetting may also occur in FL. Therefore, they propose FedCurv, which introduces Elastic Weight Consolidation (EWC) to regularize model parameters, and thus avoid forgetting the learned information from other clients. Similarly, Yao \textit{et al.} \cite{yao2020continual} propose FedCL, which utilize a proxy dataset to construct importance weight matrix of all local models to perform EWC regularization, and thus alleviate weight divergence. Dinh \textit{et al.} \cite{t2020personalized} propose pFedME, which introduce Moreau Envelope for regularization. It adds an L2 regularization term to the local objective to transform the entire FL into a two-level objective. The inner optimization obtains optimal personalized models while the outer one obtains the global model. 
% Comparing to personalized FL methods that directly start local training from a single global model, \cite{zhang2020personalized} propose FedFomo, where the server individually save a model for every client, and determine which models as well as their combination weight for each client in downstream communication. For scenarios with Non-IID data cross-silo FL, \cite{huang2021personalized} propose FedAMP, which can save a cloud version for each client’s personalized model. FedAMP considers the clients’ uploaded personalized model as message, introducing a attentive mechanism to pass these messages to different personalized cloud models and updating personalized cloud models with the weighted convex combination of the received messages. 
% To alleviate feature skew, \cite{li2021fedbn} propose FedBN. FedBN is similar to FedAvg, but it introduce Batch Normalization layers in the model. The clients only have to upload parameters except BN layers, and retain BN statistics locally, which reflect some features related to the client’s domain. 
{Karimireddy \textit{et al.} \cite{karimireddy2020scaffold} propose SCAFFOLD, utilizing control variables that reflect the update direction of the global model and the local models respectively, and then correct update direction of local model with a regularization term formed by their differences, which can help alleviate the performance degradation caused by data heterogeneity.} 
% Hanzely \textit{et al.} \cite{hanzely2020federated} propose a new formulation of FL, which no longer learn a single global model, but a mixture of the global model and the pure personalized models. To perform such FL formulation, this works proposes a novel randomized gradient-based method L2GD, for minimizing the loss function composed of average loss and penalty term. 
%To handle label distribution skew and quality skew simultaneously, \cite{xu2022fedcorr} propose a multi-stage FL framework FedCorr. Firstly FedCorr utilize LID (Local intrinsic dimension) to identify and current some noisy samples and further identify the noisy clients. Then, it invite some clean clients to fine-tune the model and further correct noisy samples. Finally FedCorr perform typical FL (e.g. FedAvg). 
Li \textit{et al.} \cite{li2021ditto} propose a personalized FL method Ditto, where each client perform local training on the global model and then on its personalized model, and applies regularization to the personalized model’s training to bring it closer to the global model. Besides, They also theoretically and empirically prove that Ditto has good robustness and fairness guarantee. Recently, Zhou \textit{et al.} \cite{zhou2024federated} propose FedLNL, which designs a novel method based on Bayesian Inference for updating local Noise Transition Matrix (NTM) and thus avoid too large gradients, and then apply a diversity product regularizer in local training to utilize this method. Zhi \textit{et al.} \cite{zhi2024knowledge} suggest that each client can guide its local training referring to other clients’ parameters, and thus propose a regularizer based on a relation cube to conduct layer-wise parameters coaching. Li \textit{et al.} \cite{li2024fednar} empirically observe that current FL algorithms are highly susceptible to the degree of weight decay, and propose FedNAR. FedNAR is an algorithmic plug-in which can help controlling the magnitude of weight decay.

\subsection{Architecture-Level Methods}
To address the federated heterogeneity challenge, making adjustments in the training structure is an emerging improvement. In this section, we discuss some heterogeneous methods that make adjustments on the FL architecture. Specifically, we classify these structure-level approaches into client selection, clustered FL, hierarchical FL, and decentralized FL.
\subsubsection{Client selection}
Client selection refers to the process of selecting a subset of available clients to participate in model updates during the training process. This selection can directly impact the efficiency and effectiveness of the training. {Effective client selection can address issues of data heterogeneity and constraints related to communication and computation \cite{huang2022stochastic}.}

{To alleviate the pressure of heterogeneity, Nishio \textit{et al.} \cite{nishio2019client} proposed that FedCS uses a greedy algorithm to solve the client selection problem with resource constraints, which allows the server to aggregate as many client updates as possible and accelerate the performance improvement of ML models.}
Yoshida \textit{et al}. \cite{yoshida2020hybrid} adopted a heuristic algorithm and proposed a hybrid learning mechanism called Hybrid-FL, trying to select the best set of clients to use their own data to train the model.
{Wang \textit{et al.} \cite{wang2020optimizing} introduced an experience-driven control framework, Favor, which intelligently selects client devices to participate in each round of FL to offset the bias introduced by non-IID data and accelerate the convergence speed.}
Xu \textit{et al.} \cite{xu2020client} proposes a stochastic optimization problem that jointly connects client selection and bandwidth allocation under long-term client constraints. 
Lai \textit{et al.} \cite{lai2021oort} proposed Oort to improve the performance of joint training and testing by guiding participant selection. Oort uses existing data to improve model accuracy and speed up training. 
Wu \textit{et al.} \cite{wu2022node} proposed a probabilistic node selection framework, FedPNS, which dynamically changes the probability of each node being selected based on the output of the optimal aggregation. 
Cho \textit{et al.} \cite{cho2022towards} presents a convergence analysis of FL of biased customer selection via efficient selection aggregation and quantifies how bias affects the convergence speed. 
Li \textit{et al.} \cite{li2022pyramidfl} proposed PyramidFL to speed up FL training while achieving higher final model performance. At the heart of PyramidFL is fine-grained client selection, not only focusing on the differences in client selection between those selected and unselected participants, but also fully exploiting data and system heterogeneity within selected clients.
Ribero \textit{et al.} \cite{ribero2022federated} proposes a federated averaging algorithm assisted by an adaptive sampling technique, F3AST, an unbiased algorithm that dynamically learns an availability-dependent client selection strategy that asymptotically minimizes client sampling variance to global model convergence The impact improves the performance of joint learning. 
\subsubsection{Clustered Federated Learning}
{Clustering FL (CFL) is used to deal with situations where the data distribution is highly heterogeneous. This approach trains the model in a more homogeneous group by clustering clients based on the similarity of their data.}
{CFL \cite{sattler2020cfl} is a method of recursively clustering clients by exploiting the cosine similarity between client gradient updates during each round of FL training. This clustering allows for a more targeted and efficient distribution of training tasks among clients with similar characteristics.} 

IFCA \cite{ghosh2022ifca} is another clustered FL method that utilizes the loss of models within different clusters on the client's local dataset to cluster clients. In each round, IFCA randomly selects a portion of clients within each cluster for training and updates the intra-cluster models by aggregating the local models within the cluster.
PACFL \cite{vahidian2023pacfl} clusters clients by uploading data information from each client to the server and taking measurements of the principal angles between client data subspaces before the FL process begins. During each round of joint learning, PACFL updates each intra-cluster model with a weighted aggregation of the local models of the intra-cluster clients, enabling personalized learning while still taking advantage of clustering.
FedSoft \cite{ruan2022fedsoft} designed to train both local personalized models and high-quality cluster models. To limit the workload of the clients, FedSoft employs proximal updating, which allows only a subset of clients to complete an optimization task in each round of communication. 
{ACFL \cite{lu2023auction} introduces a mean-shift clustering algorithm that intelligently groups clients based on their local data distributions. This clustering approach allows FL systems to identify groups of clients with similar data characteristics, enabling more targeted and efficient model training.}

\subsubsection{Hierarchical Federated Learning}
Hierarchical Federated Learning is a more organizationally complex FL architecture that employs a multilevel federation model to manage and train clients distributed at different levels. This model is usually suitable for scenarios with a multi-layered network structure. It is worth noting that hierarchical FL focuses on how to organize and optimize the learning process in networks with a well-defined hierarchical structure, e.g., federated edge learning. In contrast, clustered FL focuses more on grouping clients based on data similarity without the need for an explicit hierarchical structure.

Liu \textit{et al.} \cite{liu2020client} propose a client-edge cloud hierarchical FL system, HierFAVG, an algorithm that allows multiple edge servers to perform partial model aggregation, which allows for better communication and utility tradeoffs. 
Abad \textit{et al.} \cite{abad2020hierarchical} collaboratively learn global models by sharing local updates of model parameters instead of their dataset, and in this way use a hierarchical joint learning scheme to significantly reduce communication latency without sacrificing accuracy. 
Briggs \textit{et al.} \cite{briggs2020federated} separated clusters of clients by introducing a hierarchical clustering step that uses the similarity of the client's local updates to the global joint model. 
Xu \textit{et al.} \cite{xu2021adaptive} consider a hierarchical FL system and propose a joint problem of edge aggregation interval control and resource allocation to minimize a weighted sum of training loss and training delay. 
Lim \textit{et al.} \cite{lim2021dynamic} use an evolutionary game to model workers' edge association strategies and propose a hierarchical game framework to study the dynamics of edge association and resource allocation in self-organizing HFL networks.
{Abdellatif \textit{et al.} \cite{abdellatif2022communication} investigated the potential of hierarchical FL in IoT heterogeneous systems and proposed a user assignment and resource allocation method on a hierarchical FL architecture for IoT heterogeneous systems.} 
Zhou \textit{et al.} \cite{zhou2023hierarchical} proposed a unique clustering-based approach for participant selection using social context data. By creating different groups of edge participants and performing group-specific joint learning, the models of various edge groups are further aggregated to enhance the robustness of the global model.
Ma \textit{et al.} \cite{ma2024feduc} proposed FedUC, which clusters workers through a hierarchical aggregation structure to solve the edge heterogeneity problem and reduce the communication consumption of FL. 
Deng \textit{et al.} \cite{deng2024communication} proposed a communication cost minimization problem in the HFL framework by making decisions on edge enhancer selection and node edge association in order to achieve the target accuracy of the total communication cost required for model learning.

\subsubsection{Decentralized Federated Learning}
Decentralized federated learning (DFL) is an approach that does not rely on a central server to manage model training. Compared with traditional centralized FL, decentralized methods provide higher privacy protection.

{Li \textit{et al.} \cite{li2020blockchain} developed a decentralized FL framework BFLC based on blockchain with committee consensus. It uses blockchain for global model storage and local model update exchange \cite{che2022decentralized}, and reduces malicious client attacks through committee verification composed of honest clients.} 
Lim \textit{et al.} \cite{lim2021decentralized} proposed a deep learning-based auction mechanism to derive the valuation of the services of each cluster leader. 
Ye \textit{et al.} \cite{ye2022decentralized} proposed a robust decentralized stochastic gradient descent federated method, called soft DSGD, to solve the problem of unreliable transmission process. 
Pappas \textit{et al.} \cite{pappas2021ipls} designed a fully decentralized FL framework based in part on the Inter Planetary File System (IPFS). By using IPLS and connecting to the corresponding private IPFS network, any party can initiate the training process of a ML model, or join an ongoing training process initiated by the other party. 
Liu \textit{et al.} \cite{liu2022decentralized} proposed a general DFL framework that regularly implements multiple local updates and multiple inter-node communications to strike a balance between communication efficiency and model consistency. 
Kalra \textit{et al.} \cite{kalra2023decentralized} proposed an efficient communication scheme for decentralized FL, called ProxyFL. Each participant in ProxyFL maintains two models, one is a private model and the other is a public shared proxy model. Designed to protect participant privacy. The agent model allows efficient information exchange between participants without the need for a centralized server. 
HegedHus \textit{et al.} \cite{hegedHus2021decentralized} believe that Gossip learning is a decentralized alternative to FL, so they designed a DFL training algorithm based on the Gossip protocol. 
Qu \textit{et al.} \cite{qu2021decentralized} proposed a new architecture called DFL for drone networks (DFL-UN), which implements FL in drone networks without the need for a central entity. 
Beltran \textit{et al.} \cite{beltran2024fedstellar} introduced a novel platform, Fedstar, which allows users to create federations by customizing parameters. Fedstar aims to train FL models in different federations of physical or virtual devices in a decentralized, semi-decentralized and centralized manner. 
Zhou \textit{et al.} \cite{zhou2024defta} proposed a decentralized federated trusted average (DeFTA), as a plug-and-play replacement for FedAvg, brings immediate improvements to security, scalability, and fault tolerance in the FL process.
Cheng \textit{et al.} \cite{cheng2024deta} used a decentralized and trusted aggregation strategy to introduce DeTA, a decentralized FL system architecture, and designed a defense-in-depth model.

\subsubsection{Asynchronous Federated Learning}
Traditional FL typically employs a synchronous aggregation strategy. When devices are heterogeneous, resource utilization is constrained because the aggregation process in each training round must wait for the slower devices. {Asynchronous FL \cite{xu2023asynchronous} is a distributed ML approach designed to address the inefficiencies in training caused by disparities in computational power and communication latency among different participants in traditional FL.} In asynchronous FL, participants are not required to synchronize their model updates \cite{liu2024fedasmu}. Instead, they can independently conduct local training and update their models based on their computational speed and network conditions. The central server continuously receives updates from various participants and integrates these updates into the global model.

{Chen \textit{et al.} \cite{chen2020vafl} allow each client to run the stochastic gradient algorithm without coordinating with other clients, and thus is suitable for intermittent connections of clients. The method further uses a new perturbed local embedding technique to ensure data privacy and improve communication efficiency.}
Sprague \textit{et al.} \cite{sprague2018asynchronous} proposed a new asynchronous FL algorithm and investigated the speed of convergence when distributed over multiple edge devices under hard data constraints, and applied asynchronous FL to challenging geospatial applications.
{Nguyen \textit{et al.} \cite{nguyen2022federated} address the inability of synchronous FL to efficiently scale to hundreds of parallel client trainings by proposing a new buffered asynchronous aggregation method, FedBuff, which is independent of the choice of optimizer and combines the best features of both synchronous and asynchronous FL.}
Chen \textit{et al.} \cite{chen2020asynchronous} proposed an asynchronous online FL (ASO Fed) framework in which edge devices use continuously streaming local data for online learning and a central server aggregates model parameters from clients.
Wang \textit{et al.} \cite{wang2022asynchronous} propose a new asynchronous FL framework that designs a new centralized fusion algorithm for determining the fusion weights during global updates by taking into account the possible delays in training and uploading local models.
To solve the problems of drastically reduced training accuracy and slow convergence in asynchronous FL, Zhang \textit{et al.} \cite{zhang2023timelyfl} proposed TimelyFL, a heterogeneous perceptual asynchronous FL framework with adaptive partial training.

\section{Privacy Protection in Heterogeneous Federated Learning}

\begin{figure*}[t]
    \vspace{-1cm}
    \centering
    \includegraphics[width=0.8\textwidth]{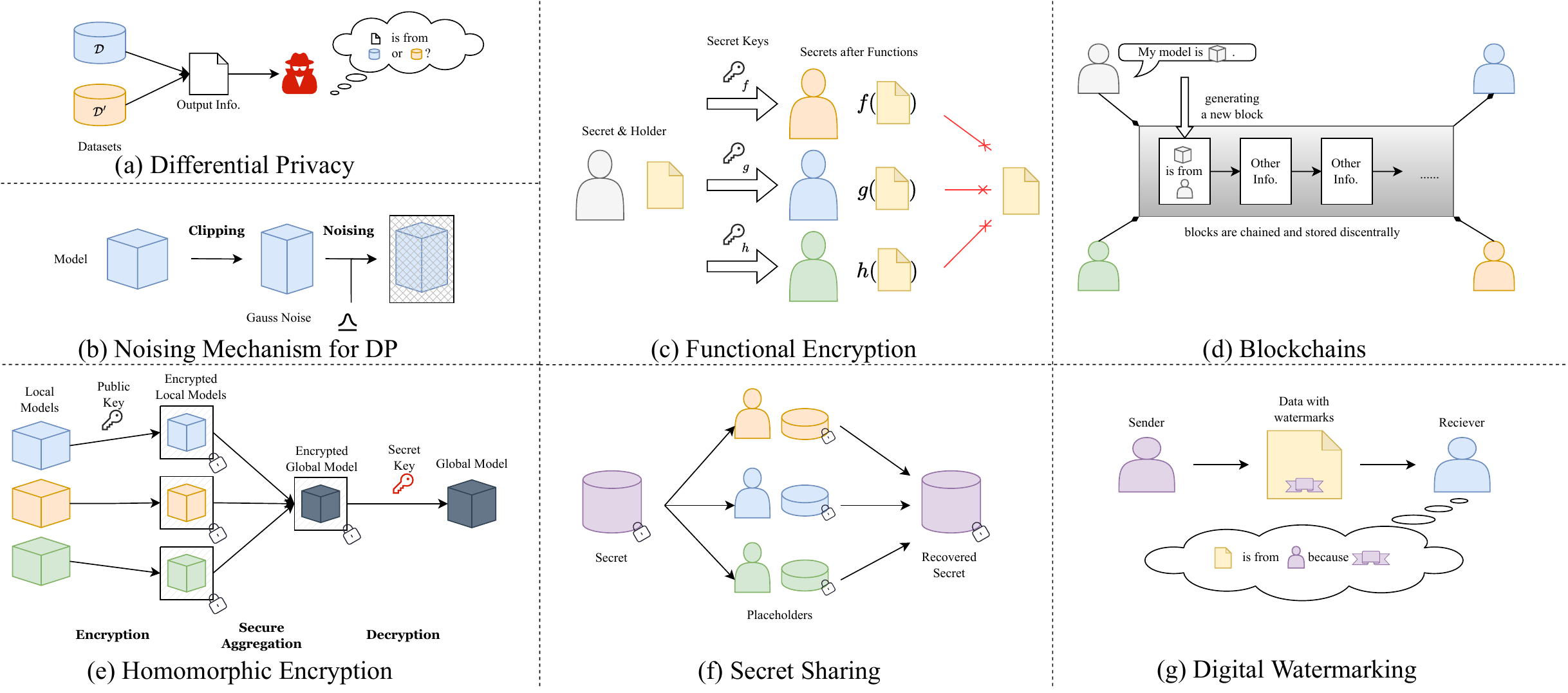}
    \caption{Illustration for Protect Methods}
    \label{fig:P}
\end{figure*}

In this section, we will focus on the privacy and security issues in heterogeneous FL.
As is well recognized, FL frameworks are deployed over the internet.
The complex and unpredictable network environment poses potential vulnerabilities
for attacks on the FL process.
This underscores the necessity of employing protective measures within FL applications.
To address the security and privacy issuses, protective methods are particularly critical.
Through years of research,
model protection can be categorized into three types \cite{lyu2022privacy}:
differential privacy, secure multi-party computation, and model tracing,
each providing pre-emptive or post-emptive protection
for the transmitted models from different perspectives.
We show common FL privacy-preserving strategies in Fig. \ref{fig:P}.

\subsection{Potential Threats}

Any stage of FL exposed in a network environment is susceptible to security and privacy threats.
We can categorize these threats
based on the methods and objectives of the attacks as follows.

\subsubsection{Byzantine Attacks}

With such attacks, malicious nodes exhibit arbitrary behavior during the training process to achieve certain malicious objectives.
The following attack methods are most representitive.

\begin{itemize}
	\item \textbf{Gradient Poisoning Attacks} \cite{shejwalkar2021manipulating}: The attacker manipulates the training direction by uploading carefully crafted malicious gradients.
	\item \textbf{Gradient Reversal Attacks} \cite{bagdasaryan2020backdoor}: The attacker reverses the local gradients before uploading, causing the model to learn in the wrong direction.
	\item \textbf{Replaying Attacks} \cite{tolpegin2020data}: The attacker captures and replays legitimate communication messages (such as local gradients) to disrupt the learning process.
	\item \textbf{Sybil Attacks} \cite{baruch2019little}: The attacker creates a large number of fake nodes to hijack the consensus in FL, amplifying the impact of the aforementioned attacks.
\end{itemize}

Furthermore, the aggregator in FL could also be a Byzantine node, with the capability to directly manipulate the training process. In such cases, stronger security measures and robust authorization methods are required.
\subsubsection{Interference Attacks}

Adversaries employing such attacks do not interfere with the training process but instead eavesdrop on the information transmitted between nodes, attempting to infer the embedded private information. 
There are also several well-known inference attacks:
% Similarly, several well-known inference attacks are also presented in the list.

\begin{itemize}

	\item \textbf{Membership Inference Attacks} \cite{nasr2019comprehensive}:
	An external eavesdropper intercepts the information transmitted
	between nodes and attempts to infer embedded private details.
% such as the participation of specific training nodes or whether certain data has been involved in the training process.
	\item \textbf{Attribute Inference Attacks} \cite{luo2021feature}:
	The attacker aims to infer certain attributes embedded in the training data,
	even if these attributes were not explicitly used in the training process.
	By analyzing the model's output or behavior, the attacker can infer specific attributes.
	\item \textbf{Model Inversion Attacks} \cite{zhu2019deep}:
	The attacker attempts to reconstruct the client's private data
	from the captured information, and it is even possible to fully reconstructbthe private dataset from the model with the assistant of GAN. 
 % \cite{yue2023gradient}

\end{itemize}

\color{black}
\subsubsection{Challanges from Heterogeneity}
The attacks mentioned before challenge the robustness of FL systems. 
Even worse, the heterogenous scenarios make it more difficult to detect or defend against such attacks.
Data heterogeneity causes the privacy information of individual clients more susceptible to inteference attacks and that
malicious Byzantine gradients are more challenging to identify \cite{chen2023guardhfl}.
Communication and device heterogeneity indicate that some communication links in FL systems are more vulnerable to attacks, primarily depending on
the protective strategies adopted by the devices.
To solve these issues, researchers propose place their interest on heterogeneity-independent protective methods,
which is able to be flexibly configured as plugins within FL systems,
such as the DP and SMC techniques discussed later.
\subsubsection{Compromise on Performence}
Even though protection methods can mitigate the adverse effects of the aforementioned attacks,
the associated costs persist.
Model perturbation-based methods enable relatively quick protection but reduce model accuracy,
whereas cryptographic approaches maintain model performance nearly unaffected \cite{mcmahan2017learning}.
However, their computational and communication overhead remains prohibitively high
in distributed environments \cite{zhang2020batchcrypt}.
Existing theory has demonstrated that in FL, it is impossible to simultaneously optimize privacy, utility, and efficiency; a trade-off is inevitable \cite{zhang2022no}. Therefore, the choice of protection methods introduced later should be guided by the preferences of administrators or users in FL systems.

\subsection{Differential Privacy}
Differential privacy (DP) \cite{dwork2006differential}, initially proposed by Dwork et al., is a privacy definition from the field of database security. Since the introduction of FL, differential privacy has attracted the interest of many researchers due to its provable and quantifiable privacy. Generally speaking, FL frameworks meet the requirements of DP through a noise mechanism \cite{dwork2014algorithmic}, which means that almost all DP methods will sacrifice model utility to obtain privacy, with little impact on training efficiency.

\subsubsection{Definitions}

Before introducing the noise mechanism in detail, we first give the formal definition of DP.

\begin{definition}[ \((\varepsilon, \delta)\)-DP]
	\label{def:dp}
	A randomized mechanism \(\mathcal{M}\) with range \(\cal{R}\) satisfies \((\varepsilon, \delta)\)-DP if,
	for any two adjacent datasets \(\cal{D},\cal{D'}\) and for any subsets of outputs \(\cal{S} \subseteq \cal{R}\) it holds that
	\begin{equation}
		\label{eqn:dp}
		\Pr [\mathcal{M} (\cal{D}) \in \mathcal{S}] \leq \exp(\varepsilon)\cdot \Pr[\mathcal{M} (\cal{D'}) \in \mathcal{S}] + \delta
	\end{equation}
	where $(\varepsilon, \delta)$ is privacy budget.
	The smaller the value of $\varepsilon$,
	the stronger the level of protection provided by differential privacy.
\end{definition}

Definition \ref{def:dp} represents the privacy of a FL algorithm by conducting
the probability of an attacker interferencing successfully whether a member is in the dataset during training.
In the context of FL, adjacent datasets is ones that differ only in the data held by a single client,
where such DP is called \textit{user-level}.

\subsubsection{Model Perturbation}

Based on relevant derivations, we generally use a noise mechanism to perturb the transmitted model, thereby making the entire FL mechanism satisfy DP. This mechanism is also known as \textbf{model perturbation}, which has two stages.
The first stage is model clipping, which involves setting a threshold $C$.
If the norm of the model exceeds this threshold,
the model is scaled proportionally until its norm equals the threshold.
The second stage involves noising,
specifically generating Gaussian noise with a
chosen standard deviation $\sigma$ to perturb the model.

\begin{equation}
	\label{eqn:mechanism}
	\begin{aligned}
		\text{Clipping:} \quad \hat{\theta} &\gets  \frac{\theta}{\max\{1, \left\| \theta \right\| / C \}} \\
		\text{Noising:} \quad \tilde{\theta} &\gets \hat{\theta} + n \sim \mathcal{N}(0, \sigma I)
	\end{aligned}
\end{equation}

The relationship between the noise amplitude $\sigma$ and the privacy budget $\varepsilon$ can be concluded from relevant DP theories,
such as Advanced Composition Theorem \cite{dwork2014algorithmic},
Moment Accountant \cite{abadi2016deep} and variant DPs like RDP \cite{mironov2017renyi} or zCDP \cite{bun2016concentrated}.
All the theories mentioned before persue a smaller upper bound for the privacy budget,
meaning that they can derive a smaller $\varepsilon$ from the same noise amplitude.

There are numerous federated learning frameworks that apply model perturbation. For instance, DP-FedAvg \cite{mcmahan2017learning}, which was initially proposed, first adopted user-level DP to protect client data. NbAFL \cite{wei2020federated} considered the noise magnitude required for differential privacy protection of model information in both upload and download links. Seif et al. \cite{seif2020wireless} consider the environment of wireless communication and propose how to implement differential privacy protection in this environment by utilizing the characteristics of wireless communication.

\subsubsection{DP with Heterogeneity}

In real-world environments, federated learning frameworks with Differential Privacy may encounter several issues due to heterogeneity. We categorize these issues as follows:

\begin{itemize}
    \item {Different clients have varying requirements for privacy, making it difficult to determine the privacy of the entire system. Such phenomenon can be called \textit{privacy heterogeneity}.}
    \item {Data heterogeneity itself can lead to a loss in model utility, and DP can make this loss more apparent \cite{xiong2021privacy}.}
\end{itemize}

In response to these issues, the academic community has proposed some solutions. PFA \cite{liu2021projected} classifies clients into public and private categories based on their privacy budgets, and then utilize the different roles for better accuracy. In some cases, the network environment where the federated learning algorithm is deployed also has heterogeneity, and FedSeC \cite{gao2022fedsec} has proposed improvements from this perspective. There are also works like DP-SCAFFOLD \cite{noble2022differentially}, which advance from the perspective of reducing model utility loss.

\subsection{Secure Multi-party Computation}

Secure Multi-party Computation (SMC) \cite{goldreich1998secure} refers to methods that ensure the security of distributed computation in complex network environments, often with security guarantees provided by cryptography. Due to the nature of cryptography \cite{menezes2018handbook}, the model information processed by SMC is almost undamaged, ensuring that the utility of the model does not decrease. However, these methods often involve complex computational processes \cite{evans2018pragmatic}, leading to significant time and space overhead during the model training process.
\subsubsection{Homomorphic Encryption}

Homomorphic encryption (HE) \cite{rivest1978data} is a specialized cryptographic scheme that imposes additional requirements on certain computational properties to be preserved beyond traditional encryption/decryption functionalities. Assuming that $+,\cdot$ are addition and multiplication on plaintext space and $\oplus, \odot$ are ones on cipher space, and $\mathsf{E}$ refers to encryption operation.
\begin{itemize}
	\item \textbf{Additive Homomorphism.} \(\mathsf{E}(a + b) = \mathsf{E}(a) \oplus \mathsf{E}(b)\) for any plaintexts $a,b$;
	\item \textbf{Multiplicative Homomorphism.} \(\mathsf{E}(a \cdot b) = \mathsf{E}(a) \odot \mathsf{E}(b)\) for any plaintexts $a,b$;
\end{itemize}

An HE scheme satisfies at least one of the properties.
When HE is employed in federated learning \cite{aono2017privacy},
the model can be encrypted before transmission,
preventing even observers or servers from accessing model information.
Building upon this foundation, leveraging the properties of homomorphic encryption,
servers can perform aggregation operations on the model without acquiring the model information.

In the current HE, there are two schemes are useful, Paillier cryptosystem \cite{paillier1999public},
and CKKS cryptosystem \cite{cheon2017homomorphic}.
The former encrypts integers and remain additional homomorphism and
the latter enables direct encryption of real number vectors, and perform homomorphic addition and multiplication,
making it more applicable to machine learning scenarios.

\subsubsection{Functional Encryption}

In conventional cryptographic schemes, possession of the private key implies the ability to fully decrypt the ciphertext. However, functional encryption (FE) \cite{boneh2011functional} imposes a limit on the amount of information that a key can access. In FE, a key \(\mathsf{sk}_f\) corresponds to a function \(f\). Assuming there is a plaintext \(x\), an entity possessing \(\mathsf{sk}_{f}\) can only obtain \(f(x)\) after decryption and cannot know the information of \(x\) (unless \(f(x)=x\) happens to be true). In a heterogeneous environment, FE requires a TTP to distribute various private keys \(\mathsf{sk}\), which implies that the security of the entire system depends on the trustworthiness of the TTP.

In federated learning, the function $f$ corresponding to the key is often set as Multi-Input Functional Encryption \cite{goldwasser2014multi}, as in HybridAlpha \cite{xu2019hybridalpha}. This setting can effectively prevent inference attacks. In response to the security issues of the TTP, Chotard et al. \cite{chotard2018decentralized} proposed a decentralized scheme, DMCFE, which allows the system to operate without any TTP, thereby enhancing security.

\subsubsection{Secrecy Sharing}

Secret sharing (SS) \cite{shamir1979share} is a technique for splitting and storing secret information, which can effectively prevent the leakage of secrets due to collusion among multiple participants. The core idea is to split the secret into $n$ parts, and at least $t$ slices are needed to fully restore the secret. This scheme is called $(t,n)$-secret sharing.
The most classic in SS is the Shamir scheme, which is built upon the Lagrange interpolation method. The role of the Shamir scheme in federated learning has been discovered by some researchers, such as the SecAgg \cite{bonawitz2017practical} and SecAgg+ \cite{bell2020secure} schemes.

\subsubsection{SMC with Heterogeneiry}
{Although SMC can provide a high level of security to FL,
it is still negatively affected by the heterogeneity of FL nodes.
The biggest challenge lies in how different clients can
generate the same key in heterogeneous networks.}

\textbf{Trustworthy Third Party (TTP)} \cite{canetti2000security}.
Clearly, clients can jointly trust an external node as a TTP,
which generates and securely distributes the key to each client.
This process is known as key distribution. 
Introducing a TTP to overcome heterogeneity is a simple method
that requires minimal computational overhead.
However, it relies on complete trust in the TTP.
If the TTP itself is malicious, the protection provided by SMC will be compromised.

\textbf{Key Agreement Protocols} \cite{diffie2022new}.
In this protocol, the key is not centrally generated;
instead, each participating node exchanges the necessary information to independently compute the same session key.
Similar to common encryption schemes,
key agreement protocols are implemented through mathematically challenging problems,
ensuring that adversaries, even upon intercepting the communication,
cannot infer the session key.
However, as a trade-off for security,
this method incurs a significant increase in computational overhead.

Both of the aforementioned solutions have their respective advantages and disadvantages, necessitating that managers select based on the specific circumstances of the system. Generally, TTP emphasizes efficiency, whereas the key agreement protocol places a greater emphasis on security.

\subsection{Model Tracing}
In contrast to the methods previously introduced, the methods discussed in this section primarily serve the function of post-hoc tracing. That is, once an attack has been detected, these methods allow for the identification and verification of the source of the attack. While these methods cannot prevent an attack from occurring in advance, they can obtain evidence after the fact, providing administrators or users of the federated learning system with an advantage at the level of rules or law.
Most model tracing methods are transparent to FL systems, functioning as plugins that seamlessly integrate into the training process while avoiding the limitations posed by the heterogeneous environment.

\subsubsection{Digital watermarking}

Digital watermarking \cite{cox2002digital} is a technique that uses algorithms to add identifiable symbols to target data without affecting the usability of the original information, effectively tracing the source of the data. In the context of federated learning, digital watermarks can be added to the transmitted model information. Even if the information is stolen or leaked, the digital watermark can be used to track or forensically investigate this violation. Some work has already combined digital watermarking technology in FL \cite{tekgul2021waffle}, and corresponding improvements have been proposed for issues such as compression \cite{nie2024fedcrmw} or robustness \cite{han2021application}.

\subsubsection{Blockchains}
Blockchain \cite{nakamoto2008bitcoin} is a distributed ledger that uses cryptography to provide security guarantees, offering immutable information recording capabilities in insecure environments, initially proposed in Bitcoin . Due to its strong security and traceability, blockchain has been extensively applied in areas such as cryptocurrencies and Non-Fungible Tokens (NFTs) \cite{wang2021non}. An important derivative application of blockchain is smart contracts, initially proposed by Ethereum. Smart contracts allow the inclusion of code that can be automatically triggered for execution in the blockchain, and the execution of this code is recorded and verified by the blockchain. In the context of federated learning, the role of blockchain is particularly evident due to the following functionalities:

\begin{itemize}
	\item \textbf{Malicious Client Identification.} Blockchain can record the information uploaded by each client, and this information is immutable, which can be used for evidence collection after an attack is discovered.
	\item \textbf{Supervision on Server.} Blockchain can also supervise the behavior of servers in scenarios where clients do not trust the servers, checking whether each aggregation behavior is malicious and constitutes an attack.
	\item \textbf{Incentive Mechanism.} Blockchain can incentivize clients based on the quality of the model they upload, in the form of blockchain cryptocurrencies, which can prevent free-riding attacks, motivate clients to collect better data for training, and ensure a certain level of fairness.
\end{itemize}

Several researchers have proposed numerous federated learning frameworks that incorporate blockchain. Among them, CREAT \cite{cui2020creat} uses smart contracts on the blockchain to verify the reliability of model information uploaded by clients. BFLC \cite{li2020blockchain}, based on blockchain, implements a decentralized federated learning committee mechanism that can be used to resist malicious attacks. Beyond theoretical research, open-source frameworks like VeryFL \cite{li2023veryfl} are advancing this field from a practical perspective.

\section{Future Research Directions}
With the rapid development of big data and artificial intelligence technologies, FL, as a privacy-preserving distributed machine learning paradigm, is becoming an important tool for processing distributed data. In heterogeneous scenarios, FL faces multiple heterogeneous challenges. 
% With the rapid development of big data and artificial intelligence technologies, people are paying increasing attention to privacy data containing a large amount of potential value. As a distributed ML paradigm that protects privacy, FL is gradually becoming an important tool for processing distributed data. In heterogeneous scenarios, FL faces multiple challenges, including data heterogeneity, model heterogeneity, task heterogeneity, communication heterogeneity, device heterogeneity, and privacy protection defects. 
Despite the existence of the above-mentioned research methods, there are still some challenges that have not been resolved. Previous solutions usually suffer from the lack of granularity of heterogeneous alignment granularity, the need for additional computational and communication costs, the relaxation of privacy-preserving assumptions. Furthermore, there is a lack of consideration of special settings such as heterogeneous models, heterogeneous devices, fairness, and so on. Therefore, in the following, we will summarize some potential research directions as follows:

\textbf{Heterogeneity}: (1) Data Distribution \cite{huang2021personalized}: Data heterogeneity is one of the main issues faced by FL in heterogeneous scenarios. Future research can start from the data itself, focusing on preprocessing data across clients and designing more intelligent data selection and filtering mechanisms to improve data quality and model training efficiency. At the same time, we can consider data generalization theory to build a bridge between the source domain and the target domain, thus alleviating the problem of data heterogeneity. 
(2) Models \cite{tan2022fedproto}: Model heterogeneity often leads to differences in model structure, parameter scale, and so on. Future research can focus on model integration and fusion techniques, unifying the semantic information of models to ensure that multiple models with different structures have the same semantic branches, supporting low-loss splicing and fusion of models with different structures. Additionally, for model heterogeneity issues, we can explore optimization techniques such as adaptive training adjustments and model pruning to improve the training efficiency and performance of the model.
(3) Tasks \cite{jiang2020customized}: Future research can focus on the analysis and utilization of task relevance, improving the generalization ability of the model by exploring the inherent connections between tasks. Furthermore, we can explore dynamic adjustment and adaptation techniques for model tasks to meet the needs of different tasks and improve the adaptability of the model, even to be able to handle federated requirements with completely different tasks. 
(4) Communication \cite{zheng2023federated}: FL relies on the robustness of communication. Studying adaptive network topologies can help mitigate the impact of communication heterogeneity on the central server and improve the overall performance of FL. Additionally, we can focus on the design of fault-tolerant mechanisms for FL to mitigate the negative impact of clients being offline on the overall system. Strategies such as optimizing data transmission formats, reducing communication rounds, and designing efficient communication scheduling strategies are also viable solutions.
(5) Devices \cite{xu2021optimizing}: Device heterogeneity is mainly reflected in differences in the types of devices participating in FL, computing capabilities, storage capacity, and other aspects. Studying the partitioning and distribution strategies of model parameters and techniques such as model compression and quantization will be feasible solutions to address device heterogeneity issues.

\textbf{Federated Fairness}: In heterogeneous scenarios, differences in data distribution, computing resources, and other factors among different participants may lead to fairness issues in the FL process \cite{ezzeldin2023fairfed}. Future research can focus on improving the fairness of FL algorithms to ensure that different participants have fair opportunities and benefits during model training. Additionally, we can establish fairness evaluation indicators and mechanisms to quantitatively evaluate and optimize the fairness of FL algorithms. 

\textbf{Secure and Trustworthy Strategies}: Security and trustworthiness are important prerequisites for the widespread application of FL \cite{wang2021federated}. Future research can focus on the design and implementation of secure and trustworthy strategies, including participant identity verification, communication encryption, and model update verification. Additionally, we can explore secure and trustworthy solutions based on advanced cryptography techniques such as zero-knowledge proofs and homomorphic encryption to improve the overall security performance of FL.

\textbf{Incentive Mechanisms}: In heterogeneous scenarios, due to differences in the interests and contributions of different participants, it is necessary to design reasonable incentive mechanisms to promote their active participation and cooperation \cite{wang2023incentive}. Future research can focus on the design and optimization of incentive mechanisms, including reward allocation strategies and contribution evaluation methods. Additionally, we can explore incentive mechanism designs based on methods such as game theory and auction theory to achieve more fair and effective resource allocation and interest coordination.

% \textbf{Transfer Learning}: Transfer learning can transfer knowledge from one domain to another, thereby accelerating model training in new domains \cite{feng2022semi}. In federated scenarios, data may be generated in real-time. Therefore, studying how to use FL to achieve cross-domain transfer learning has important practical significance for improving the adaptability and performance of FL.

\textbf{Foundation Models}: Foundation models have demonstrated their potential in a variety of scenarios. However, there are potential privacy concerns due to the large amount of data that needs to be introduced during the training process. FL as a privacy-preserving approach, it would be meaningful work to think about the combination with the foundation model. This combination can be considered from a variety of perspectives, for instance, thinking about the federated data component for the foundation models \cite{charles2024towards}. 

\textbf{Plug and Play Edge Participants:} with the development of mobile technology and IoT devices, federated learning participants may not be online all the time due to power and network limitations. The flexibility of the devices makes the joining and dropping of clients frequent. Therefore, it is necessary to investigate federated learning algorithms with flexibility and scalability.

In conclusion, there are numerous potential research directions in FL, especially in heterogeneous scenarios. Future research should focus on addressing these challenges and exploring new techniques to improve the performance, efficiency, and fairness of FL.

% \section{Conclusion}
% In this paper, we provide insights into key concepts in the field of heterogeneous federated learning and the multiple challenges it faces. In particular, we analyze in detail the heterogeneity of data, models, tasks, communications, and devices, which pose numerous algorithmic challenges in federated learning systems. To address these challenges, in this paper, we systematically categorize existing research approaches by classifying them into data-level, model-level, and structure-level according to the processing hierarchy, each of which corresponds to a specific problem and solution strategy.
% Our categorization approach provides a clear perspective for understanding the efficacy of various types of algorithms in coping with heterogeneous environments. With this hierarchical strategy, researchers and practitioners can more efficiently select methods that fit their specific needs. 
% Finally, this paper looks at future research directions in heterogeneous federated learning, which will undoubtedly continue to be a dynamic and challenging research direction in the field of ML as technology advances and application needs increase. We look forward to more innovative research results in this field in the future to promote the development of technology and popularization of applications.

% use section* for acknowledgment
%	\section*{Acknowledgment}
%	
%	The research is supported .

\bibliographystyle{IEEEtran}
\bibliography{references}

\end{document}